\documentclass[10pt,journal,compsoc]{IEEEtran}
\usepackage{multibib}
\newcommand{\para}[1]{\vspace{0.5\baselineskip}\noindent\textbf{{#1}.~}}
\usepackage{amsmath}
\usepackage{makecell}
\usepackage{graphbox}
\usepackage{arydshln}
\usepackage[bookmarks,backref=true,linkcolor=black]{hyperref}
\hypersetup{
	pdfauthor = {},
	pdftitle = {},
	pdfsubject = {},
	pdfkeywords = {},
	colorlinks=true,
	linkcolor= black,
	citecolor= black,
	pageanchor=true,
	urlcolor = black,
	plainpages = false,
	linktocpage
}

\usepackage[ruled,linesnumbered]{algorithm2e}
\usepackage{footmisc}
\usepackage{multirow}
\usepackage{amsfonts}
\usepackage{xcolor}
\usepackage{enumitem}
\usepackage{microtype}

\newcommand{\newpos}[1]{\widehat{#1}}
\newcommand{\pcaorder}[1]{\sigma(#1)}
\newcommand{\pcaaxis}{\mathbf{a}}
\DeclareMathOperator*{\proj}{proj}
\DeclareMathOperator*{\diag}{diag}
\newcommand{\SOG}{\mathcal{R}}
\DeclareMathOperator*{\argmin}{arg\,min}

\newcommand{\alignweight}[1]{w_{#1}}
\newcommand{\regweight}[1]{r_{#1}}
\newcommand{\numnodes}{|\mathcal{V}_{\mathcal{G}}|}
\newcommand{\initpar}[1]{\overline{#1}}
\newcommand{\lowerpar}[1]{\underline{#1}}

\newcommand{\numsourcept}{|\mathcal{V}|}
\newcommand{\numtargetpt}{|\mathcal{U}|}

\ifCLASSOPTIONcompsoc
  \usepackage[nocompress]{cite}
\else
  \usepackage{cite}
\fi

\ifCLASSINFOpdf
\else

\fi

\begin{document}

\title{Fast and Robust Non-Rigid Registration Using Accelerated Majorization-Minimization}

\author{

	Yuxin Yao,~
	Bailin Deng,~\IEEEmembership{Member,~IEEE},~
	Weiwei Xu,~\IEEEmembership{Member,~IEEE,}~
	Juyong Zhang$^\dagger$,~\IEEEmembership{Member,~IEEE}	
	\IEEEcompsocitemizethanks{\IEEEcompsocthanksitem Y. Yao and J. Zhang are with School of Mathematical Sciences,
		University of Science and Technology of China.



		\IEEEcompsocthanksitem B. Deng is with School of Computer Science and Informatics, Cardiff University.
		\IEEEcompsocthanksitem W. Xu is with State Key Lab of CAD $\&$ CG, Department of Computer science, Zhejiang University.}

	\thanks{$^\dagger$Corresponding author. Email: \texttt{juyong@ustc.edu.cn}.}

}

\markboth{~}%
{Yao \MakeLowercase{\textit{et al.}}: Fast and Robust Non-Rigid Registration}

\IEEEtitleabstractindextext{%
\begin{abstract}
Non-rigid 3D registration, which deforms a source 3D shape in a non-rigid way to align with a target 3D shape, is a classical problem in computer vision. Such problems can be challenging because of imperfect data (noise, outliers and partial overlap) and high degrees of freedom. Existing methods typically adopt the $\ell_{p}$ type robust norm to measure the alignment error and regularize the smoothness of deformation, and use a proximal algorithm to solve the resulting non-smooth optimization problem. However, the slow convergence of such algorithms limits their wide applications. In this paper, we propose a formulation for robust non-rigid registration based on a globally smooth robust norm for alignment and regularization, which can effectively handle outliers and partial overlaps. The problem is solved using the majorization-minimization algorithm, which reduces each iteration to a convex quadratic problem with a closed-form solution. 
We further apply Anderson acceleration to speed up the convergence of the solver, enabling the solver to run efficiently on devices with limited compute capability. Extensive experiments demonstrate the effectiveness of our method for non-rigid alignment between two shapes with outliers and partial overlaps, with quantitative evaluation showing that it outperforms state-of-the-art methods in terms of registration accuracy and computational speed. The source code is available at \url{https://github.com/yaoyx689/AMM_NRR}.
\end{abstract}

}

\maketitle

\IEEEdisplaynontitleabstractindextext

\IEEEpeerreviewmaketitle

\section{Introduction}
\IEEEPARstart{W}{ith}
the popularity of depth acquisition devices such as Kinect, PrimeSense and  depth sensors on smartphones, techniques for 3D object tracking and reconstruction from point clouds have enabled various applications. Non-rigid registration is a fundamental problem for such techniques, especially for reconstructing dynamic objects.
Since depth maps obtained from structured light or time-of-flight cameras often contain outliers and holes, a robust non-rigid registration algorithm is needed to handle such data. Moreover, real-time applications require high computational efficiency for non-rigid registration.

Given a source surface and a target surface, each represented as a point cloud or a mesh,
non-rigid registration aims to find a deformation field for the source surface to align it with the target surface.
This problem is typically solved via optimization.
The objective energy often includes alignment terms that measure the deviation between the target surface and deformed source surface, as well as regularization terms that enforce the smoothness of the deformation field.
Many existing methods formulate these terms using the $\ell_2$-norm, which penalizes alignment and smoothness errors across the whole surface~\cite{amberg2007optimal,li2008global,li2009robust}. On the other hand, the ground-truth alignment may induce large errors for these terms in some local regions due to noise, outliers, partial overlaps, or articulated motions. 
The $\ell_2$ formulations can inhibit such large localized errors and lead to erroneous alignment.
To improve the alignment accuracy, recent works have utilized sparsity-promoting norms for these terms, such as the $\ell_1$-norm~\cite{yang2015sparse,li2018robust,jiang2019huber} and the $\ell_0$-norm~\cite{guo2015robust}.
The sparsity optimization enforces small error values on most parts of the surface while allowing for large errors in some local regions, improving the robustness of the registration process.
However, the resulting optimization problem can be non-smooth and more challenging to solve.
Existing methods often use proximal algorithms such as the alternating direction method of multipliers (ADMM), which can suffer from slow convergence to high-accuracy solutions~\cite{boyd2011distributed}.

In this paper, we propose a new approach to robust non-rigid registration with fast convergence. The key idea is to enforce sparsity using Welsch's function~\cite{holland1977robust}, which has been utilized for robust filtering of images~\cite{ham2015robust} and meshes~\cite{zhang2018static}, as well as robust rigid registration~\cite{zhang2021fast}.
We formulate an optimization that applies Welsch's function to both the alignment term and the regularization term.
Unlike the $\ell_p$-norms, Welsch's function is globally smooth and avoids non-smooth optimization.
We solve the optimization problem using the majorization-minimization (MM) algorithm~\cite{Lange2016mm}. It iteratively constructs a surrogate function for the objective energy based on the current variable values and minimizes the surrogate function to update the variables, and is guaranteed to converge to a local minimum.
This leads to a convex quadratic problem in each iteration, which can be solved efficiently in closed form.
To speed up the convergence of the MM algorithm, we regard it as a fixed-point iteration and apply Anderson acceleration~\cite{anderson1965iterative,walker2011anderson}, a well-established technique to accelerate the convergence of fixed-point iterations. 
Experimental results verify the robustness of our method as well as its superior performance compared to existing robust registration approaches.

In summary, our main contributions include:
\begin{itemize}[leftmargin=*]
	\item We formulate an optimization problem for non-rigid registration, using Welsch's function to induce sparsity for alignment error and transformation smoothness. The proposed formulation effectively improves the robustness and accuracy of the results.
	\item We propose an MM algorithm to solve the optimization problem, which only involves minimizing a convex quadratic surrogate function in each iteration; we also apply Anderson acceleration to speed up its convergence. The simple form of the surrogate function, together with the effectiveness of Anderson acceleration, helps to improve the computational efficiency of robust non-rigid registration compared to existing approaches.
\end{itemize} 

A preliminary version of this work appeared in~\cite{yao2020quasi}. The main additions in the current version include: (1) a more robust method for the deformation graph construction; (2) a more efficient optimization solver that combines the MM algorithm with Anderson acceleration; (3) a more extensive evaluation of the performance.
\section{Related work}
Non-rigid registration has been extensively studied in computer vision, medical image processing and computer graphics. The reader can refer to the surveys in~\cite{Tam2013Registration} and \cite{Deng2022} for a more complete overview of past research. 
In the following, we focus on works that are closely related to our method. 

\para{Non-rigid Registration Models}
Various optimization-based methods perform non-rigid registration by minimizing the alignment error, often in conjunction with other constraints. 
Extending the classical iterative closest point (ICP) algorithm for rigid registration,  Amberg et al.~\cite{amberg2007optimal} proposed a non-rigid ICP algorithm that incrementally deforms the source model to align with the target model. 
Li et al.~\cite{li2008global} adopted an embedded deformation approach~\cite{sumner2007embedded} to express a non-rigid deformation using a deformation graph, and simultaneously optimized the correspondence between source and target scans, the confidence weights for the correspondence, and a warping field that aligns the source with the target. 
In addition, various geometric constraint terms have been introduced to preserve the local shape of the source model during deformation. Examples include preservation of Laplacian operators~\cite{wand2007reconstruction,liao2009modeling, huang2011global}, as-conformal-as-possible deformations~\cite{wu2019global}, and as-rigid-as-possible deformations~\cite{yang2019global}.

Other methods tackle the registration problem from a statistical perspective.  
By treating point cloud alignment as a probability density estimation problem, Myronenko et al.~\cite{Andriy2010Point} proposed the Coherent Point Drift (CPD) algorithm, which encourages displacement vectors to point in similar directions to improve the coherence of the transformation.
Hirose~\cite{hirose2020bayesian} reformulated CPD in a Bayesian setting to improve robustness with a better guarantee of convergence.
Hontani et al.~\cite{hontani2012robust} incorporated a statistical shape model and a noise model into the non-rigid ICP framework, and detected outliers based on their sparsity. 
Jian et al.~\cite{jian2005robust, Jian2011robust} represented each point set as a mixture of Gaussians and treated point set registration as a problem of aligning two mixtures. 
Also with a statistical framework, Wand et al.~\cite{Wand2009Efficient} used a meshless deformation model to perform the pairwise alignment. 
Ma et al.~\cite{ma2015robust} proposed an ${L_2}E$ estimator to build more reliable sparse and dense correspondences. These methods typically require calculations for each possible pair of points from the source and the target, which is time-consuming for problems with a large number of points.

Some dynamic reconstruction and tracking frameworks involve non-rigid registration between a template model and the live frames. The registration is performed by alternately updating the correspondences and solving the transformation on the GPU to produce real-time results~\cite{zollhofer2014real, newcombe2015dynamicfusion, Yu17bodyfusion,  yu2018doublefusion, li2020robust}.  For such applications, the non-rigid registration tasks are usually less challenging due to the small deviation between the source and target models, and real-time performance is favored over high-precision solutions. 

When the transformation between the source and target models is nearly isometric, some approaches utilize the geodesic distance of the surface as an invariant metric to directly optimize the global correspondences~\cite{bronstein2006generalized,dubrovina2011approximately,chen2015robust}. However, these methods may not be suitable for noisy mesh surfaces, as the noise can distort the geodesic distance and lead to erroneous alignment.

With the growing popularity of deep learning in computer graphics and vision tasks, it has also been adopted for non-rigid registration problems. 
Shimada et al.~\cite{shimada2019dispvoxnets} represented the source and target models using regular voxel grids, and learned the displacement field using datasets of deformable objects with known correspondence.
For non-rigid 3D reconstruction, Bo{\v{z}}i{\v{c}} et al.~\cite{bozic2020deepdeform} extracted features from the input RGB-D frames and matched them to obtain sparse correspondences, and used these initial correspondences to improve the accuracy of registration. For non-rigid tracking, Li et al.~\cite{li2020learning} proposed an alignment term based on deep features learned through CNN to improve the robustness of the optimization formulation, and introduced a learning-based preconditioner to speed up the convergence of the PCG method used in the optimization solver.
Bo{\v{z}}i{\v{c}} et al.~\cite{bovzivc2020neural} combined a dense correspondences prediction module based on an optical flow network of RGB images, and a differentiable alignment module to achieve more accurate real-time non-rigid tracking.  These methods improve the robustness and/or speed compared with optimization-based methods, by training on carefully constructed datasets. Feng et al.~\cite{feng2021recurrent} represented the non-rigid transformation as a point-wise combination of several rigid transformations. They proposed an unsupervised framework to learn the alignment, using a differentiable loss that measures similarity between 3D shapes via their depth images from multiple views. However, this method cannot handle partial overlaps.

\para{Robust Metrics} Since the data generated from a 3D acquisition process often contains noise, outliers and partial overlaps, various robust metrics have been applied in optimization-based approaches to better handle such cases. Unlike the classical $\ell_2$ error metrics that require the error to be small across the whole shape, such robust metrics promote the sparsity of the error and allow for large errors in some local regions to accommodate data imperfections.
Yang et al.~\cite{yang2015sparse} introduced an $\ell_1$ term for the smoothness of the transformation, which is suitable for models with articulated motions. 
Li et al.~\cite{li2018robust} additionally utilized an $\ell_1$-norm for the alignment error, to improve robustness against noises and outliers. 
Guo et al.~\cite{guo2015robust} proposed an $\ell_0$-based motion regularizer for robust motion tracking and surface reconstruction.
Zampogiannis et al.~\cite{zampogiannis2019topology} applied the Huber-$L_1$ loss to the difference between the transformation parameters at neighboring deformation graph nodes, to promote piecewise smooth deformation fields.
Other works used the Geman-McClure function for the non-rigid alignment term~\cite{yu2018doublefusion,xu2019unstructuredfusion,su2020robustfusion}. 
In this paper, we adopt Welsch's function for robust non-rigid registration, which results in a smooth optimization problem that allows for an efficient numerical solver.

\para{Anderson Acceleration} Anderson acceleration~\cite{anderson1965iterative} was initially designed for the iterative solution of nonlinear integral equations. 
In recent years, it has been shown effective in accelerating the convergence of fixed-point iterations~\cite{walker2011anderson,toth2015convergence,Evans2020}.
In each iteration, Anderson acceleration utilizes $m$ previous iterates to construct a new iterate that converges faster, which can be considered as a quasi-Newton method for finding a root of the residual function~\cite{Fang2009}.
It has been adopted to accelerate first-order solvers in computer graphics~\cite{peng2018anderson,Zhang2019ADMM, Ouyang2020}, as well as ICP-based methods for rigid registration~\cite{artem2018aaicp,zhang2021fast}. 
In this paper, we apply Anderson acceleration to speed up the convergence of our non-rigid registration solver.

\section{Model}
Given a source surface and a target surface, non-rigid registration computes a deformation field on the source surface to align it with the target surface. 
We assume that the source surface is represented with a set of sample points 
$\mathcal{V} = \{\mathbf{v}_1, \ldots, \mathbf{v}_{\numsourcept}\in \mathbb{R}^3\}$ connected by a set of edges $\mathcal{E}$ that define their neighboring relations, and $\mathcal{U} = \{\mathbf{u}_1, \ldots, \mathbf{u}_{\numtargetpt} \in \mathbb{R}^3\}$ is a set of sample points on the target surface. 
In this paper, we focus on the case where the source surface is a triangle mesh, so that $\mathcal{V}$ and $\mathcal{E}$ are the mesh vertices and edges, respectively. However, the formulation is also applicable when the input surface is a point cloud and $\mathcal{V}$ are the points: in this case, the edge set $\mathcal{E}$ can be constructed by connecting each point with its nearest neighbors. Following~\cite{sumner2007embedded}, we define the deformation field using an embedded deformation graph $\mathcal{G} = (\mathcal{V}_{\mathcal{G}}, \mathcal{E}_{\mathcal{G}})$, where $\mathcal{V}_{\mathcal{G}} = \{\mathbf{p}_1,  \ldots, \mathbf{p}_{\numnodes}\in \mathbb{R}^3\} \subset \mathcal{V}$ is the set of graph nodes that lie on the source surface, and $\mathcal{E}_{\mathcal{G}}$ is the set of edges. Each graph node is associated with an affine transformation $\mathbf{X}_j = [\mathbf{A}_j, \mathbf{t}_j]$, where $\mathbf{A}_j\in \mathbb{R}^{3\times 3}$ and $\mathbf{t}_j\in \mathbb{R}^3$. In the following, we use 
$$\mathbf{X} = [\mathbf{X}_1, \mathbf{X}_2, \ldots, \mathbf{X}_{\numnodes}] \in \mathbb{R}^{4 \numnodes \times 3}$$ 
to denote the concatenation of all $\mathbf{X}_j$. Using the deformation graph, each sample point $\mathbf{v}_i$ on the source surface is influenced by a set $\mathcal{I}(\mathbf{v}_i)$ of nearby deformation graph nodes:
\begin{equation}
\mathcal{I}(\mathbf{v}_i) = \{\mathbf{p}_j \in \mathcal{V}_{\mathcal{G}} \mid D(\mathbf{v}_i,\mathbf{p}_j) < R\},
\label{eq:InfluenceNodes}
\end{equation}
where $D(\cdot, \cdot)$ denotes the geodesic distance on the source surface, and $R$ is a given radius parameter. The new position of the sample point $\mathbf{v}_i$ is computed as
\begin{equation}
\label{Eq:deformed_position}
\newpos{\mathbf{v}}_i = \sum_{\mathbf{p}_j\in\mathcal{I}(\mathbf{v}_i)} \alignweight{ij} \cdot(\mathbf{A}_j(\mathbf{v}_i-\mathbf{p}_j)+\mathbf{p}_j+\mathbf{t}_j),
\end{equation}
where $\alignweight{ij}$ is a distance-dependent normalized weight~\cite{li2009robust}:
\begin{equation}
\alignweight{ij} = \frac{(1 - D^2(\mathbf{v}_i, \mathbf{p}_j)/R^2)^3}{\sum_{\mathbf{p}_k\in\mathcal{I}(\mathbf{v}_i)}(1 - D^2(\mathbf{v}_i, \mathbf{p}_k)/R^2)^3}. 
\label{Eq:weight_data}
\end{equation}
From the given source and target surfaces, we first construct a deformation graph, and then optimize the transformations associated with the graph nodes to deform the source surface and align it with the target surface. Compared with existing non-rigid registration methods that define the transformation field on each source vertex such as~\cite{li2018robust}, the use of a deformation graph can effectively reduce the number of variables for the optimization problem and significantly reduce the computational time.
The remainder of this section explains each step of our methods in detail.

\subsection{Construction of deformation graph}
\begin{figure}[htb]
	\centering
	\includegraphics[width=\columnwidth]{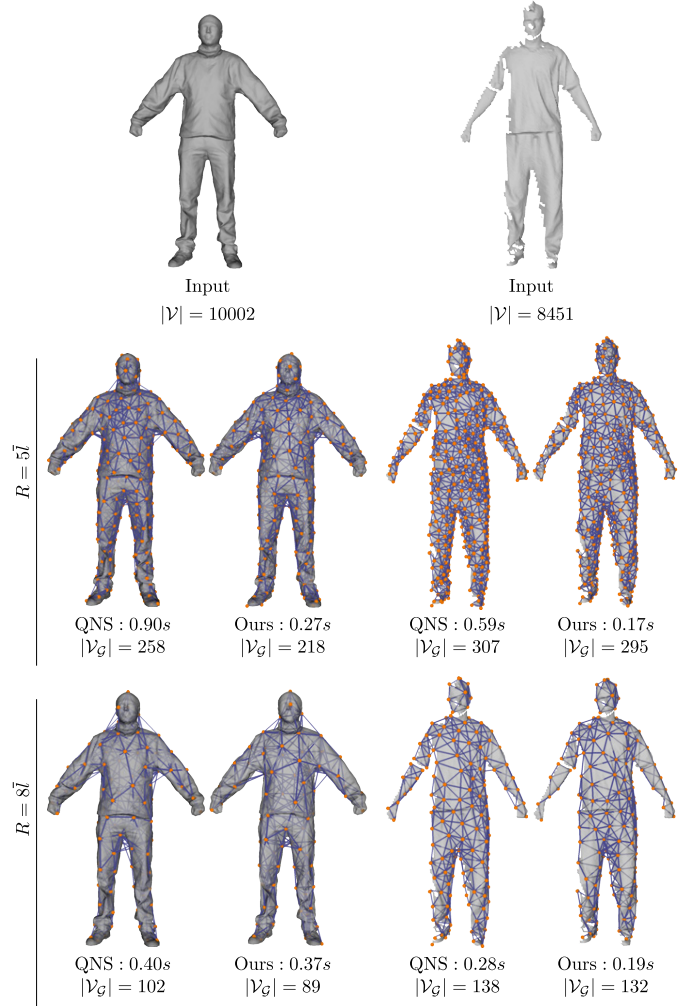}
	\caption{Comparison of deformation graph construction by~\cite{yao2020quasi} and our method on ``crane'' (left) and ``march2'' (right) datasets of~\cite{vlasic2008articulated}.}
	\label{Fig:deformation_graph}
\end{figure}

To avoid excessive degrees of freedom for the deformation, we would like to reduce the number of deformation graph nodes while ensuring that for each source point $\mathbf{v}_i$ the set of influencing nodes $\mathcal{I}(\mathbf{v}_i)$ (defined in Eq.~\eqref{eq:InfluenceNodes}) is not empty. To this end, we simultaneously construct the set $\mathcal{V}_{\mathcal{G}}$ of deformation graph nodes and the set $\mathcal{I}(\mathbf{v}_i)$ of influencing nodes for each point as follows.
We first initialize $\mathcal{V}_{\mathcal{G}}$ and all $\mathcal{I}(\mathbf{v}_i)$ to empty sets.
Then we perform PCA on all source points $\{\mathbf{v}_i\}$, and sort them based on their projections onto the axis corresponding to the largest eigenvalue of the covariance matrix. 
We go through all the points according to this order, and add a source point $\mathbf{v}_j$ to the deformation graph nodes $\mathcal{V}_{\mathcal{G}}$ if either $\mathcal{V}_{\mathcal{G}}$ or $\mathcal{I}(\mathbf{v}_j)$ is empty.
Once a point $\mathbf{v}_j$ is added to the deformation graph nodes, we also add $\mathbf{v}_j$ to the influencing node set for any point $\mathbf{v}_i$ that is within geodesic distance $R$ from $\mathbf{v}_j$ (i.e., $D(\mathbf{v}_i, \mathbf{v}_j) < R$).
To locate all such points efficiently, we first start from $\mathbf{v}_j$ and perform a breadth-first search on the surface to identify the largest neighborhood $\mathcal{B}(\mathbf{v}_j)$ of points whose Euclidean distance to $\mathbf{v}_j$ is smaller than $R$. Since the Euclidean distance is no larger than the geodesic distance, any point with a geodesic distance smaller than $R$ must belong to $\mathcal{B}(\mathbf{v}_j)$. Hence we compute the geodesic distance from $\mathbf{v}_j$ to each point $\mathbf{q} \in \mathcal{B}(\mathbf{v}_j)$, and add $\mathbf{v}_j$ to $\mathcal{I}(\mathbf{q})$ if $D(\mathbf{v}_j, \mathbf{q}) < R$.  
To compute the geodesic distance, we construct a sub-mesh consisting of all faces on the source surface mesh that are incident with the vertices from $\mathcal{B}(\mathbf{v}_j)$, and apply the exact geodesic algorithm of~\cite{Qin2016VTP} (VTP) on this sub-mesh. We use VTP for the computation because it can produce accurate results while being efficient on small meshes.
After the set of deformation graph nodes is constructed, we connect two nodes with an edge if there exists a point that is influenced by both of them. 
Algorithm~\ref{Alg:construct_graph} summarizes our method for constructing the deformation graph.
Note that in the preliminary version~\cite{yao2020quasi} of this paper, the deformation graph construction method needs to compute the geodesic distance from each graph node to all the source points; in comparison, our new method only requires the geodesic distance to the source points within the sphere centered at the graph node and with a radius $R$, which involves a significantly smaller number of geodesic distance values. This allows us to compute the exact geodesic distance using VTP to achieve more robust results, whereas the method in~\cite{yao2020quasi} needs to use the fast marching method (FMM) to compute an approximate geodesic distance to avoid excessive computational costs.  
Fig.~\ref{Fig:deformation_graph} shows some examples of deformation graphs constructed using our new method and the method from~\cite{yao2020quasi} and their computational time, where the FMM geodesic distance is computed using an open-source implementation\footnote{\url{https://github.com/sywe1/geodesic-computation}}. We can see that our method is notably faster when using the same radius parameter.

\textbf{\emph{Remark.}} An alternative way of constructing the deformation graph nodes is to repeatedly sample the source points and add them to the graph using farthest point sampling~\cite{Moenning2003Fast}, until the geodesic distance between graph nodes and the farthest point is smaller than the radius parameter $R$ or the number of graph nodes reaches a certain threshold. However, this would require computing geodesic distance between faraway points on the surface, which tends to be much slower than our approach where the geodesic distance is only evaluated between nearby points.  

\begin{algorithm}[t]
	\KwIn{The source surface mesh with vertex set $\{\mathcal{V}\}$; the radius parameter $R$. 
	}
	\KwResult{The deformation graph $\{\mathcal{V}_{\mathcal{G}}, \mathcal{E}_{\mathcal{G}}\}$;
	The set of influencing nodes $\mathcal{I}(\mathbf{v}_i)$ for each $\mathbf{v}_i \in \mathcal{V}$. 
	} 
	\BlankLine
	Initialize $\mathcal{V}_{\mathcal{G}} = \emptyset$, $\mathcal{E}_{\mathcal{G}} = \emptyset$, and  $\mathcal{I}(\mathbf{v}_i) = \emptyset$ for all $\mathbf{v}_i \in \mathcal{V}$ \;
	Perform PCA on all points in $\mathcal{V}$ to obtain the axis $\pcaaxis{}$ corresponding to the largest eigenvalue of the covariance matrix\;
	Let $\mathbf{v}_{\pcaorder{1}},\mathbf{v}_{\pcaorder{2}},\ldots \mathbf{v}_{\pcaorder{|\mathcal{V}|}}$ be the ordering of points in $\mathcal{V}$ according to their projections to $\pcaaxis{}$\;
	
	\tcp{Construct deformation graph nodes and influencing node sets}
	\For{$i = 1, 2 , \ldots, |\mathcal{V}|$}{
	    \If{$\mathcal{V}_{\mathcal{G}}$ is empty OR  $\mathcal{I}(\mathbf{v}_{\pcaorder{i}})$ is empty}
	    {
	        Add $\mathbf{v}_{\pcaorder{i}}$ to $\mathcal{V}_{\mathcal{G}}$\;
	        Perform breadth-first search from $\mathbf{v}_{\pcaorder{i}}$ to find the largest neighborhood $\mathcal{B}(\mathbf{v}_{\pcaorder{i}})$ where 
	        $\|\mathbf{v}_{\pcaorder{i}} - \mathbf{q}\| < R ~\forall \mathbf{q} \in \mathcal{B}(\mathbf{v}_{\pcaorder{i}})$\;
	        Use VTP~\cite{Qin2016VTP} to compute the geodesic distance $D(\mathbf{v}_{\pcaorder{i}}, \mathbf{q})$ for all $\mathbf{q} \in \mathcal{B}(\mathbf{v}_{\pcaorder{i}})$\;
	        \For{each $\mathbf{q} \in \mathcal{B}(\mathbf{v}_{\pcaorder{i}})$}
	        {
	            \If{$D(\mathbf{v}_{\pcaorder{i}}, \mathbf{q}) < R$}
	            {
	                Add $\mathbf{v}_{\pcaorder{i}}$ to $\mathcal{I}(\mathbf{q})$\;
	            }
	        }
	    }
	}
	
	\tcp{Construct deformation graph edges}
	\For{$\mathbf{v}_i\in \mathcal{V}$}
	{
	    \For{any $\mathbf{p}_j, \mathbf{p}_k \in \mathcal{I}(\mathbf{v}_i)$ with $j < k$}
	    {
	        Add $(\mathbf{p}_j, \mathbf{p}_k)$ to $\mathcal{E}_{\mathcal{G}}$\;
	    }
	}			
	\caption{Construction of the deformation graph and the influencing node sets.}
	\label{Alg:construct_graph}
\end{algorithm}

\subsection{Problem formulation}
Similar to existing non-rigid registration methods, we would like to align the source and target surfaces using a smooth transformation field that is locally close to rigid deformation.
Therefore, we minimize the following target function to determine the transformation $\mathbf{X}$ that is defined over the deformation graph nodes:
\begin{equation}
\label{Eq:objective_func}
E(\mathbf{X}) = E_{\text{align}}(\mathbf{X}) + \alpha E_{\text{reg}}(\mathbf{X}) + \beta E_{\text{rot}}(\mathbf{X}).
\end{equation}
Here $E_{\text{align}}$ measures the alignment error,  $E_{\text{reg}}$ is a smoothness regularization term for the transformation field, and $E_{\text{rot}}$ penalizes the deviation between the transformation $\mathbf{X}_j$ on each node and a rigid transformation.
$\alpha$ and $\beta$ are user-specified weights. The definition for each term is explained below.

\begin{figure}[t!]
	\centering
	\includegraphics[width=\columnwidth]{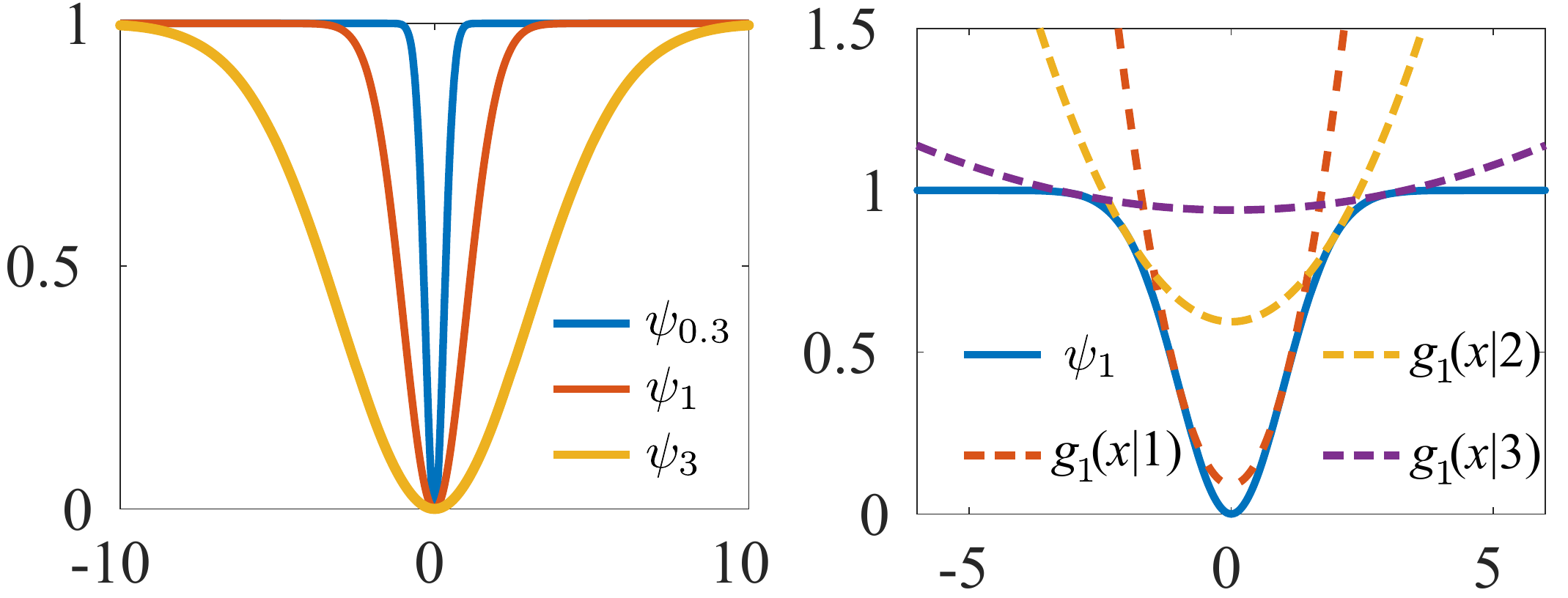}
	\caption{Left: function $\psi_{\nu}$ with different values of the parameter $\nu$. Right: different surrogate functions for the function $\psi_{\nu}$ with $\nu=1$.}
	\label{Fig:welsch_sur}
\end{figure}

\para{Alignment term} For each transformed source point $\newpos{\mathbf{v}}_i$, we find its closest point $\mathbf{u}_{\rho(i)}$ on the target surface as the corresponding point:
\[
\mathbf{u}_{\rho(i)} := \argmin_{\mathbf{u} \in \mathcal{U}} \| \mathbf{u} - \newpos{\mathbf{v}}_i  \|.
\]
Traditional methods such as~\cite{amberg2007optimal} measure the alignment error using the sum of squared distance from all the transformed source points to their corresponding points. However, outliers and partial overlaps can result in large distances between some point pairs for the ground-truth transformation, which can lead to a large value of such $\ell_2$ alignment measures and prevent the optimization from finding the ground-truth alignment. Recently, robust metrics such as the $\ell_p$-norm ($p \leq 1$)~\cite{bouaziz2013sparse,yang2015sparse,li2018robust,guo2015robust} have been utilized instead for the alignment error.
It is less sensitive to noises and partial overlaps, since such $\ell_p$-norm minimization allows for large distances at some points.
However, due to the non-smoothness of such $\ell_p$-norm formulations, their    numerical minimization can be much more expensive than the $\ell_2$-norm.
For example, the problem is solved in~\cite{bouaziz2013sparse} with an iterative algorithm that alternately updates the point correspondence and the transformation; the transformation update step minimizes a target function that involves the $\ell_p$-norm and is performed using an ADMM solver, which is time-consuming and lacks a convergence guarantee due to the non-convexity of the problem. 
In this paper, inspired by recent works in robust image filtering~\cite{ham2015robust} and mesh processing~\cite{zhang2018static}, we adopt Welsch's function~\cite{holland1977robust} as a robust metric and derive the following alignment error term:
\begin{equation}
E_{\text{align}}(\mathbf{X}) = \sum_{\mathbf{v}_i\in\mathcal{V}}\psi_{\nu_a}(\|\newpos{\mathbf{v}}_i - \mathbf{u}_{\rho(i)}\|),
\label{eq:AlignmentTerm}
\end{equation}
where $\nu_a > 0$ is a user-specified parameter, and
\begin{equation}
\label{Eq:Welsch_func}
\psi_{\nu}(x) =1 - \exp(-\frac{x^2}{2\nu^2}).
\end{equation}
As shown in Fig.~\ref{Fig:welsch_sur}, the function $\psi_{\nu}$ is monotonically increasing on $[0, +\infty)$ while being bounded from above by 1. In addition, when $\nu_a \to 0$, the error measure in Eq.~\eqref{eq:AlignmentTerm} approaches the $\ell_0$-norm of the distance values between the corresponding points. Thus it penalizes the distance between corresponding points, without being overly sensitive to large distances caused by outliers and partial overlaps. 
Moreover, as shown later in Section~\ref{sec:optimization}, the use of Welsch's function enables us to design an efficient optimization solver with guaranteed convergence. 

\para{Regularization term} 
Ideally, the transformation induced by two neighboring deformation graph nodes $\mathbf{p}_i$ and $\mathbf{p}_j$ should be consistent on their overlapping influenced region. We measure such consistency at $\mathbf{p}_i$ using the following term:
\begin{equation}
\label{Eq:reg_diff}
\mathbf{D}_{ij} = \regweight{ij} (\mathbf{A}_j(\mathbf{p}_i - \mathbf{p}_j) + \mathbf{p}_j + \mathbf{t}_j - (\mathbf{p}_i + \mathbf{t}_i)).
\end{equation}
Here the term $\mathbf{A}_j(\mathbf{p}_i - \mathbf{p}_j) + \mathbf{p}_j + \mathbf{t}_j - (\mathbf{p}_i + \mathbf{t}_i)$ is the difference between transformations induced by $\mathbf{X}_i$ and $\mathbf{X}_j$ at $\mathbf{p}_i$, which is commonly used for the regularization of transformation fields defined on a deformation graph~\cite{sumner2007embedded,li2009robust}.
$\regweight{ij}$ is a normalization weight that accounts for the distance between $\mathbf{p}_i$ and $\mathbf{p}_j$:
\begin{equation}
\regweight{ij} = \frac{2|\mathcal{E}_{\mathcal{G}}| \cdot \|\mathbf{p}_i - \mathbf{p}_j\|^{-1}}{\sum_{\mathbf{p}_i\in\mathcal{V}_{\mathcal{G}}}\sum_{\mathbf{p}_j\in \mathcal{N}(\mathbf{p}_i)}\|\mathbf{p}_i - \mathbf{p}_j\|^{-1}},
\end{equation}
where $\mathcal{N}(\mathbf{p}_i)$ denotes the set of neighboring deformation graph nodes for the node $\mathbf{p}_i$.

For non-rigid registration, the magnitude of $\|\mathbf{D}_{ij}\|$ should be small for most parts of the deformation graph. On the other hand, the optimal transformation field may induce large magnitudes of $\|\mathbf{D}_{ij}\|$ in some local regions such as the joints of a human body due to  articulated motion. Therefore, we apply the Welsch's function to  $\|\mathbf{D}_{ij}\|$  to penalize its magnitude across the deformation graph while allowing for large magnitudes in some regions, resulting in the following regularization term:
\begin{equation}
E_{\text{reg}} = \sum_{\mathbf{p}_i\in\mathcal{V}_{\mathcal{G}}}\sum_{\mathbf{p}_j\in\mathcal{N}(\mathbf{p}_i)}\psi_{\nu_r}(\|\mathbf{D}_{ij}\|),
\label{eq:RegTerm}
\end{equation}
where $\nu_r > 0$ is a user-specified parameter.

\para{Rotation Matrix term}
To preserve local surface shapes during the deformation, we would like each transformation $ \mathbf{X}_i$ to be close to a rigid transformation. 
We enforce this property by penalizing the difference between the affine transformation matrix $\mathbf{A}_i$ and its closest projection onto the rotation matrix group $\mathcal{R} = \{\mathbf{R}\in\mathbb{R}^{3\times 3}|\mathbf{R}\mathbf{R}^T=\mathbf{I}, \det(\mathbf{R})>0\}$, and define the term $E_{\text{rot}}$ as 
\begin{equation}
E_{\text{rot}} = \sum_{\mathbf{p}_i\in\mathcal{V}_{\mathcal{G}}}\|\mathbf{A}_i-\proj\nolimits_{\mathcal{R}}(\mathbf{A}_i)\|_F^2,
\end{equation}
where $\proj\nolimits_{\mathcal{R}}(\cdot)$ is the projection operator:
\[
\proj\nolimits_{\mathcal{R}}(\mathbf{A})
 = \argmin_{\mathbf{R} \in \mathcal{R}} \|\mathbf{R} - \mathbf{A}\|.
\]

\subsection{Numerical optimization}
\label{sec:optimization}

The target function of the optimization problem~\eqref{Eq:objective_func} is non-linear and non-convex. Thanks to the use of Welsch’s function, it can be solved efficiently using the majorization-minimization (MM) algorithm~\cite{Lange2016mm}. Specifically, given the variable values $\mathbf{X}^{(k)}$ in the current iteration, the MM algorithm constructs a surrogate function ${E}(\mathbf{X}|{\mathbf{X}^{(k)}})$ for the target function $E$ such that 
\begin{equation}
\begin{aligned}
{E}({\mathbf{X}^{(k)}}|{\mathbf{X}^{(k)}}) &= E(\mathbf{X}^{(k)}),\\
{E}(\mathbf{X}|{\mathbf{X}^{(k)}}) & \geq E(\mathbf{X}) \quad \forall \mathbf{X} \neq \mathbf{X}^{(k)}.
\end{aligned}
\label{eq:MMConditions}
\end{equation}
Then the variables are updated by minimizing the surrogate function
\begin{equation}
\mathbf{X}^{(k+1)} = \argmin_{\mathbf{X}}E(\mathbf{X}|{\mathbf{X}^{(k)}}).
\label{eq:MMUpdate}
\end{equation}
This guarantees the decrease of the target function in each iteration, since Equations~\eqref{eq:MMConditions} and \eqref{eq:MMUpdate} imply that 
\[
    E(\mathbf{X}^{(k+1)}) \leq {E}(\mathbf{X}^{(k+1)}|{\mathbf{X}^{(k)}})
    \leq {E}(\mathbf{X}^{(k)}|{\mathbf{X}^{(k)}}) =  E(\mathbf{X}^{(k)}).
\]
As a result, the iterations are guaranteed to converge to a local minimum~\cite{Lange2016mm}. In comparison, existing solvers for minimizing non-convex $\ell_p$-norms such as ADMM~\cite{bouaziz2013sparse} or iteratively reweighted least squares~\cite{Daubechies2010} either lack a convergence guarantee or rely on additional strong assumptions to guarantee convergence. In the following, we explain the construction of the surrogate function and its numerical minimization.

\subsubsection{Surrogate function}
To construct the surrogate function ${E}(\mathbf{X}|{\mathbf{X}^{(k)}})$, we note that there is a convex quadratic surrogate function for the Welsch's function $\psi_{\nu}$ at $y$~\cite{ham2015robust} (see Fig.~\ref{Fig:welsch_sur} right):
\begin{equation}
    {\psi}_{\nu}(x|y) = \psi_{\nu}(y) + \frac{1 - \psi_{\nu}(y)}{ 2 \nu^{2}}\left(x^{2}-y^{2}\right).
    \label{eq:WelschSurrogateFunc}
\end{equation}
This function bounds the Welsch's function from above, and the two function graphs touch at $x = y$.
Applying this to ${E}_{\textrm{reg}}$ in Eq.~\eqref{eq:RegTerm}, we obtain a convex quadratic surrogate function 
\begin{equation}
\begin{aligned}
&{E}_{\text{reg}}(\mathbf{X}|{\mathbf{X}^{(k)}})\\
=~& \frac{1}{2\nu_r^2}\sum_{\mathbf{p}_i\in\mathcal{V}_{\mathcal{G}}}\sum_{\mathbf{p}_j\in\mathcal{N}(\mathbf{p}_i)}\exp
\left(-\frac{\|\mathbf{D}_{ij}^{(k)}\|^2}{2\nu_r^2}\right)\|\mathbf{D}_{ij}\|^2,
\end{aligned}
\label{eq:ERegSurrogate}
\end{equation}
where $\mathbf{D}_{ij}^{(k)}$ is evaluated using Eq.~\eqref{Eq:reg_diff} at $\mathbf{X}^{(k)}$. 
Note that in Eq.~\eqref{eq:ERegSurrogate} we have ignored some constant terms as they do not affect the optimization.
Similarly, we can apply Eq.~\eqref{eq:WelschSurrogateFunc} to $E_{\text{align}}$ and ignore some constant terms, to obtain a surrogate function 
\begin{equation}
\frac{1}{2\nu_a^2}\sum_{\mathbf{v}_i \in \mathcal{V}}\exp\left(-\frac{\|\hat{\mathbf{v}}_i^{(k)} - \mathbf{u}_{\rho(i)}^{(k)}\|^2}{2\nu_a^2}\right)\|\newpos{\mathbf{v}}_i - \mathbf{u}_{\rho(i)}\|^2,
\label{eq:AlignNativeSurrogate}
\end{equation}
where the point $\newpos{\mathbf{v}}_i^{(k)}$ is calculated with Eq.~\eqref{Eq:deformed_position} at $\mathbf{X}^{(k)}$, and $\mathbf{u}_{\rho(i)}^{(k)}$ is the closest point to  $\newpos{\mathbf{v}}_i^{(k)}$ on the target surface. However, this is not a quadratic function of the new position $\newpos{\mathbf{v}}_i$, since the point $\mathbf{u}_{\rho(i)}$ depends non-linearly on $\newpos{\mathbf{v}}_i$.
To obtain a more simple form, we note that the term $\|\newpos{\mathbf{v}}_i -\mathbf{u}_{\rho(i)}\|^2$ has a quadratic surrogate function $\|\newpos{\mathbf{v}}_i -\mathbf{u}_{\rho(i)}^{(k)}\|^2$ at $\mathbf{X}^{(k)}$, since by definition the closest point $\mathbf{u}_{\rho(i)}$ satisfies $\|\newpos{\mathbf{v}}_i - \mathbf{u}_{\rho(i)}\|^2 \leq \|\newpos{\mathbf{v}}_i - \mathbf{u}\|$ for all $\mathbf{u} \in \mathbb{R}^3$. Applying this to Eq.~\eqref{eq:AlignNativeSurrogate}, we obtain the following convex quadratic surrogate function for $E_{\text{align}}$ at $\mathbf{X}^{(k)}$ up to an additional constant:
\begin{equation}
\begin{aligned}
 & {E}_{\text{align}}(\mathbf{X}|{\mathbf{X}^{(k)}})\\
 =~ & \frac{1}{2\nu_a^2}\sum_{\mathbf{v}_i \in \mathcal{V}}\exp\left(-\frac{\|\hat{\mathbf{v}}_i^{(k)} - \mathbf{u}_{\rho(i)}^{(k)}\|^2}{2\nu_a^2}\right)\|\newpos{\mathbf{v}}_i - \mathbf{u}_{\rho(i)}^{(k)}\|^2,
\end{aligned}
\label{eq:AlignQuadSurrogate}
\end{equation}
For the rotation term $E_{\text{rot}}$, we can similarly derive a quadratic surrogate function as
\begin{equation}
{E}_{\text{rot}}(\mathbf{X}|{\mathbf{X}^{(k)}}) = \sum_{\mathbf{p}_i \in \mathcal{V}_{\mathcal{G}}}\|\mathbf{A}_i-\proj\nolimits_{\SOG}(\mathbf{A}_i^{(k)})\|.
\end{equation}
In total, the surrogate function of $E(\mathbf{X})$ can be written as
\begin{equation}
\begin{aligned}
&{E}(\mathbf{X}|{\mathbf{X}^{(k)}})\\
=~ &{E}_{\text{align}}(\mathbf{X}|{\mathbf{X}^{(k)}}) + \alpha {E}_{\text{reg}}(\mathbf{X}|{\mathbf{X}^{(k)}}) + \beta {E}_{\text{rot}}(\mathbf{X}|{\mathbf{X}^{(k)}}).
\end{aligned}
\label{Eq:surrogate_func}
\end{equation} 
This surrogate function is minimized to obtain the updated variables $\mathbf{X}^{(k+1)}$.

\subsubsection{Numerical minimization}
Since ${E}(\mathbf{X}|{\mathbf{X}^{(k)}})$ is a convex quadratic function, we can minimize it by solving a linear system.
To derive the system, we first rewrite ${E}(\mathbf{X}|{\mathbf{X}^{(k)}})$ in a matrix form. The term ${E}_{\text{align}}(\mathbf{X}|{\mathbf{X}^{(k)}})$ can be written as
\begin{equation}
{E}_{\text{align}}(\mathbf{X}|{\mathbf{X}^{(k)}}) = \|\mathbf{W}_a(\mathbf{F} \mathbf{X} - \mathbf{Q})\|_F^2,
\end{equation}
where 
$\mathbf{W}_a = \diag(\sqrt{w_1^a}, \ldots , \sqrt{w_{\numsourcept}^a}) \in \mathbb{R}^{\numsourcept \times \numsourcept}$ 
with $$w_i^a = \frac{1}{2\nu_a^2}\exp(-\frac{\|\newpos{\mathbf{v}}_i^{(k)} - \mathbf{u}_{\rho(i)}^{(k)}\|^2}{2\nu_a^2}),$$ 
$\mathbf{F}$ is a block matrix  $\{\mathbf{F}_{ij}\}_{1 \leq i \leq \numsourcept \atop 1 \leq j \leq \numnodes} \in \mathbb{R}^{\numsourcept \times 4 \numnodes}$ 
with
\[
\mathbf{F}_{ij}
= \left\{
\begin{array}{ll}
\alignweight{ij} \cdot [ \mathbf{v}_i^T - \mathbf{p}_j^T, 1 ] & \textrm{if}~\mathbf{p}_j \in \mathcal{I}(\mathbf{v}_i)\\
\mathbf{0} & \textrm{otherwise}
\end{array}
\right.,
\]
and $\mathbf{Q} = [ \mathbf{Q}_1, \mathbf{Q}_2, \ldots, \mathbf{Q}_{\numsourcept} ]^T \in \mathbb{R}^{\numsourcept \times 3}$ with
\[
    \mathbf{Q}_i = \mathbf{u}_{\rho(i)}^{(k)} - \sum_{\mathbf{p}_j\in \mathcal{I}(\mathbf{v}_i)} \alignweight{ij}\mathbf{p}_j.
\]
Similarly, the term ${E}_{\text{reg}}(\mathbf{X}|{\mathbf{X}^{(k)}})$ can be written as
\begin{equation}
{E}_{\text{reg}}(\mathbf{X}|{\mathbf{X}^{(k)}}) = \|\mathbf{W}_r(\mathbf{H} \mathbf{X} -\mathbf{Y})\|_F^2,
\label{eq:regMatrixForm}
\end{equation}
where the matrices $\mathbf{H} \in \mathbb{R}^{2 |\mathcal{E}_{\mathcal{G}}| \times 4 \numnodes}$ and $\mathbf{Y} \in \mathbb{R}^{2 |\mathcal{E}_{\mathcal{G}}| \times 3}$ encode the computation of a term $\mathbf{D}_{ij}$ in the same row: the row in $\mathbf{H}$ has two non-zero blocks $[\regweight{ij}(\mathbf{p}_i^T - \mathbf{p}_j^T), \regweight{ij}]$ and $[0, 0, 0, -\regweight{ij}]$ corresponding to $\mathbf{X}_j$ and $\mathbf{X}_i$ respectively, whereas the row in $\mathbf{Y}$ has elements $[\regweight{ij}(\mathbf{p}_i^T - \mathbf{p}_j^T)]$. The diagonal matrix $\mathbf{W}_r \in \mathbb{R}^{2 |\mathcal{E}_{\mathcal{G}}| \times 2 |\mathcal{E}_{\mathcal{G}}|}$ has an element $\sqrt{\frac{1}{2 \nu_r^2}\exp\Big(-\frac{\|\mathbf{D}_{ij}^{(k)}\|^2}{2 \nu_r^2}\Big)}$ in the row corresponding to $\mathbf{D}_{ij}$.
Finally, the term ${E}_{\text{rot}}(\mathbf{X}|{\mathbf{X}^{(k)}})$ can be written as
\[
{E}_{\text{rot}}(\mathbf{X}|{\mathbf{X}^{(k)}})
= \|\mathbf{J} \mathbf{X} - \mathbf{Z}\|_F^2,
\]
where 
\begin{align*}
    \mathbf{J} & = \diag (1, 1, 1, 0, 1, 1, 1, 0, \ldots, 1, 1, 1, 0) \in \mathbb{R}^{4\numnodes \times 4\numnodes},\\
    \mathbf{Z} & = [ \proj\nolimits_{\mathcal{R}}(\mathbf{A}_1), \mathbf{0}, \ldots, \proj\nolimits_{\mathcal{R}}(\mathbf{A}_{\numnodes}), \mathbf{0} ]^T \in \mathbb{R}^{4 \numnodes \times 3}.
\end{align*}

Using the matrix forms, the updated variable $\mathbf{X}^{(k+1)}$ is computed via the linear system
\begin{equation}
\label{Eq:surr_solution}
\mathbf{K}^{(k)} \mathbf{X}^{(k+1)} = \mathbf{B}^{(k)},
\end{equation}
where 
\begin{align*}
\mathbf{K}^{(k)} & = \mathbf{F}^T\mathbf{W}_a^T \mathbf{W}_a \mathbf{F} + \alpha \mathbf{H}^T\mathbf{W}_r^T \mathbf{W}_r\mathbf{H} + \beta\mathbf{J}^T \mathbf{J},\\
\mathbf{B}^{(k)} & =  \mathbf{F}^T\mathbf{W}_a^T \mathbf{W}_a \mathbf{Q} + \alpha \mathbf{H}^T\mathbf{W}_r^T \mathbf{W}_r \mathbf{Y} + \beta\mathbf{J}^T \mathbf{J} \mathbf{Z}.
\end{align*}
We repeat this step until either the maximum number of iterations $I_{\max}$ is reached, or the deformed source model converges (indicated by the condition $\max_i\|\hat{\mathbf{v}}_i^{(k+1)}-\hat{\mathbf{v}}_i^{(k)}\|<\epsilon$ where $\epsilon$ is a user-specified threshold).

\subsubsection{Choice of parameters $\nu_a$ and $\nu_r$}
The parameters $\nu_a$ and $\nu_r$ play an important role in the robustness of our method. The surrogate functions in Eqs.~\eqref{eq:ERegSurrogate} and \eqref{eq:AlignQuadSurrogate} show that our solver actually minimizes a quadratic error function, where the weights for the alignment error terms and the regularization terms are updated dynamically as a Gaussian function of the current error norms with variance $\nu_a^2$ and $\nu_r^2$ respectively. On one hand, we would like  $\nu_a$ and $\nu_r$ to be sufficiently small in the final stage of the optimization, such that the Gaussian weights can effectively filter out the influence of alignment and rotation terms with large errors due to outliers, partial overlaps and articulated motions. On the other hand, fixing $\nu_a^2$ and $\nu_r^2$ to small values can reduce the effectiveness of the optimization, as many terms would have large errors at the beginning of the optimization and be excluded due to their small Gaussian weights.
Therefore, we gradually decrease the values of $\nu_a$ and $\nu_r$ during the optimization. Specifically, we initialize $\nu_a$ and $\nu_r$ to values $\initpar{\nu_a}$ and $\initpar{\nu_r}$ that are large enough to accommodate a large number of terms, and use these values $\nu_a, \nu_r$ to perform optimization until convergence. Then we reduce $\nu_a$ and $\nu_r$ by half, and use the previous optimized variables as initial values to continue the optimization until convergence. We repeat this process until the $\nu_a$ reaches a user-specified lower bound $\lowerpar{\nu_a}$.
We choose  $\initpar{\nu_a} = \overline{d}$ and $\lowerpar{\nu_a} = \overline{l}/\sqrt{3}$ following~\cite{zhang2021fast}, and set $\initpar{\nu_r} = 3 \overline{l}$, where $\overline{d}$ is the median value of the initial distance between each source point and its closest point on the target surface, and $\overline{l}$ denotes the average edge length of the source model.

\subsubsection{Anderson acceleration}
The MM solver monotonically decreases the target function and converges quickly to an approximate solution, but it can take a long time to converge to a high-accuracy solution. To speed up the final convergence, we adopt Anderson acceleration~\cite{walker2011anderson}, a well-established approach for accelerating fixed-point iterations. To do so, we consider our MM solver as a fixed point iteration
\begin{equation}
\mathbf{X}^{(k+1)} = G(\mathbf{X}^{(k)})
\label{eq:MMiteration}
\end{equation}
with a residual function 
\[
    F(\mathbf{X}) = G(\mathbf{X}) - \mathbf{X}.
\]
If the MM solver converges to a solution $\mathbf{X}^*$, then $\mathbf{X}^*$ is a fixed point of the iteration \eqref{eq:MMiteration} and $F(\mathbf{X}^*) = \mathbf{0}$.
In each iteration, Anderson acceleration uses the current iterate $\mathbf{X}^{(k)}$ and $m$ previous iterates $\mathbf{X}^{(k-1)}, \ldots, \mathbf{X}^{(k-m)}$ to compute an accelerated iterate $\mathbf{X}_{\text{AA}}$ as an affine combination of $G(\mathbf{X}^{(k)}), G(\mathbf{X}^{(k-1)}), \ldots, G(\mathbf{X}^{(k-m)})$:
\begin{equation}
\label{Eq:AA_solution}
\mathbf{X}_{AA} = G(\mathbf{X}^{(k)})-\sum_{j=1}^m\theta_j^*(G(\mathbf{X}^{(k-j+1)})-G(\mathbf{X}^{(k-j)})),
\end{equation}
with the coefficients $\{\theta_j^*\}$ computed by solving a linear least-squares problem:
\begin{equation*}
(\theta_1^*,...,\theta_m^*) = \argmin_{\theta_1,...,\theta_m}\|F^{(k)}-\sum_{j=1}^m\theta_j(F^{(k-j+1)}-F^{(k-j)})\|^2,
\end{equation*}
where $F^{(k)} = F(\mathbf{G}^{(k)})$.
Since the accelerated iterate $\mathbf{X}_{AA}$ may suffer from instability and lead to stagnation, we follow the stabilization approach from~\cite{peng2018anderson} and only accept it as the new iterate $\mathbf{X}^{(k+1)}$ if it decreases the target function; otherwise, we revert to the original MM update $G(\mathbf{X}^{(k)})$ which is guaranteed to decrease the target function, i.e.,
\[
    \mathbf{X}^{(k+1)}
    = \left\{
    \begin{array}{ll}
      \mathbf{X}_{AA}   & \text{if}~ E(\mathbf{X}_{AA}) < E(\mathbf{X}^{(k)}),\\
      G(\mathbf{X}^{(k)})   & \text{otherwise}.
    \end{array}
    \right.
\]
This approach guarantees a monotonic decrease of the target function.
Similar to~\cite{peng2018anderson}, some intermediate results that are computed during the evaluation of $E(\mathbf{X}_{AA})$ \textemdash{} in particular the closest points $\{\mathbf{u}_{\rho(i)}\}$ and the rotation matrices $\{\proj\nolimits_{\mathcal{R}}(\mathbf{A}_i)\}$ \textemdash{} can be reused for the MM step in the next iteration if $\mathbf{X}_{AA}$ is accepted. This helps to alleviate the computational overhead of the acceptance mechanism.
Algorithm~\ref{Alg:AA-NRR} summarizes our approach that combines the MM update with Anderson acceleration.

\textbf{\emph{Remark.}} In the preliminary version~\cite{yao2020quasi} of the current paper, the same optimization problem is solved using a different approach, which iteratively constructs a non-quadratic surrogate function and minimizes it using an inner L-BFGS solver.
One benefit of our new approach is that it involves fewer parameters for the solver. In particular, the inner L-BFGS solver in~\cite{yao2020quasi} requires a threshold parameter for the sufficient decrease condition used in its line search, as well as two parameters for its termination criteria. These parameters need to be chosen properly to achieve the best performance of the solver. In comparison, our new solver does not involve a line search or an inner solver, and hence does not require the tuning of such parameters. More numerical comparisons between the two approaches can be found in Section~\ref{sec:LBFGSComparison}.

\begin{algorithm}[t!]
	\KwIn{~ $\{\mathcal{V}, \mathcal{E}\}, \mathcal{U}$: source and target models; 
	}
	\KwResult{$\mathbf{X}^*$ converging to a local minimum of $E(\mathbf{X})$.}
	\BlankLine
	Initialize deformation graph $\{\mathcal{V}_{\mathcal{G}}, \mathcal{E}_{\mathcal{G}}\}$ by Alg.~\ref{Alg:construct_graph}. \\
	Initialize $\mathbf{X}^{(0)}$ with identity transformations\;
	$\nu_a = \overline{\nu_a}$; \quad $\nu_r = \overline{\nu_r}$;\\
	$k=1$\;
	\While{\texttt{TRUE}}{
		$k_{\text{start}}=k-1$;$~~~~E_{\textrm{prev}} = +\infty$\;
		\Repeat{$\max_i\|\hat{\mathbf{v}}_i^{(k+1)}-\hat{\mathbf{v}}_i^{(k)}\|<\epsilon$ or $(k-k_{\text{start}})==I_{\max}$}
		{
			Find corresponding point $\mathbf{u}_{\rho(i)}^{(k)}$ for each $\newpos{\mathbf{v}}_i^{(k)}$\;
			Update the projection $\proj_{\mathcal{R}}(\mathbf{A}_j^{(k)})$\;
			Calculate objective function $E$ by Eq.~\eqref{Eq:objective_func}\;
			\If{$E > E_{\textrm{prev}}$}
			{
				$\mathbf{X}^{(k)}=\mathbf{X}'$\;
				\textbf{continue}\;
			}
			$E_{\textrm{prev}} = E$\;
			Update weight matrices $\mathbf{W}_a$ and $\mathbf{W}_r$\;
			$\mathbf{X}' = \arg\min\limits_{\mathbf{X}}{E}(\mathbf{X}|{\mathbf{X}^{(k)}})$ by solving Eq.~\eqref{Eq:surr_solution}\;
			\tcp{Anderson acceleration}
			$G^{(k)} = \mathbf{X}'$; \quad $F^{(k)}=G^{(k)}-\mathbf{X}^{(k)}$\;
			$m_k = \min(k-k_{\text{start}}, m)$\;
			$(\theta_1^\ast, \ldots, \theta_{m_k}^\ast)$ =\\ 
			$\arg\min \| F^{(k)} - \sum\limits_{j=1}^{m_k} \theta_j (F^{(k-j+1)} - F^{(k-j)}) \|_F^2$\;
			$\mathbf{X}^{(k+1)}=G^{(k)} - \sum\limits_{j=1}^m \theta_j^\ast (G^{(k-j+1)} - G^{(k-j)})$\;
			$k=k+1$\;
		}
		\lIf{$\nu_a == \underline{\nu_a}$}
		{
			\Return{$\mathbf{T}^{(k)}$}	
		}
		$\nu_a=\max(\nu_a/2, \underline{\nu_a})$; \quad
		$\nu_r=\nu_r/2$\;
		$k = k+1$\;
	}		
	\caption{Robust non-rigid registration using Welsch's function and Anderson acceleration.}
	\label{Alg:AA-NRR}
\end{algorithm}

\section{Results}

\begin{figure*}[t]
	\centering
	\includegraphics[width=\textwidth]{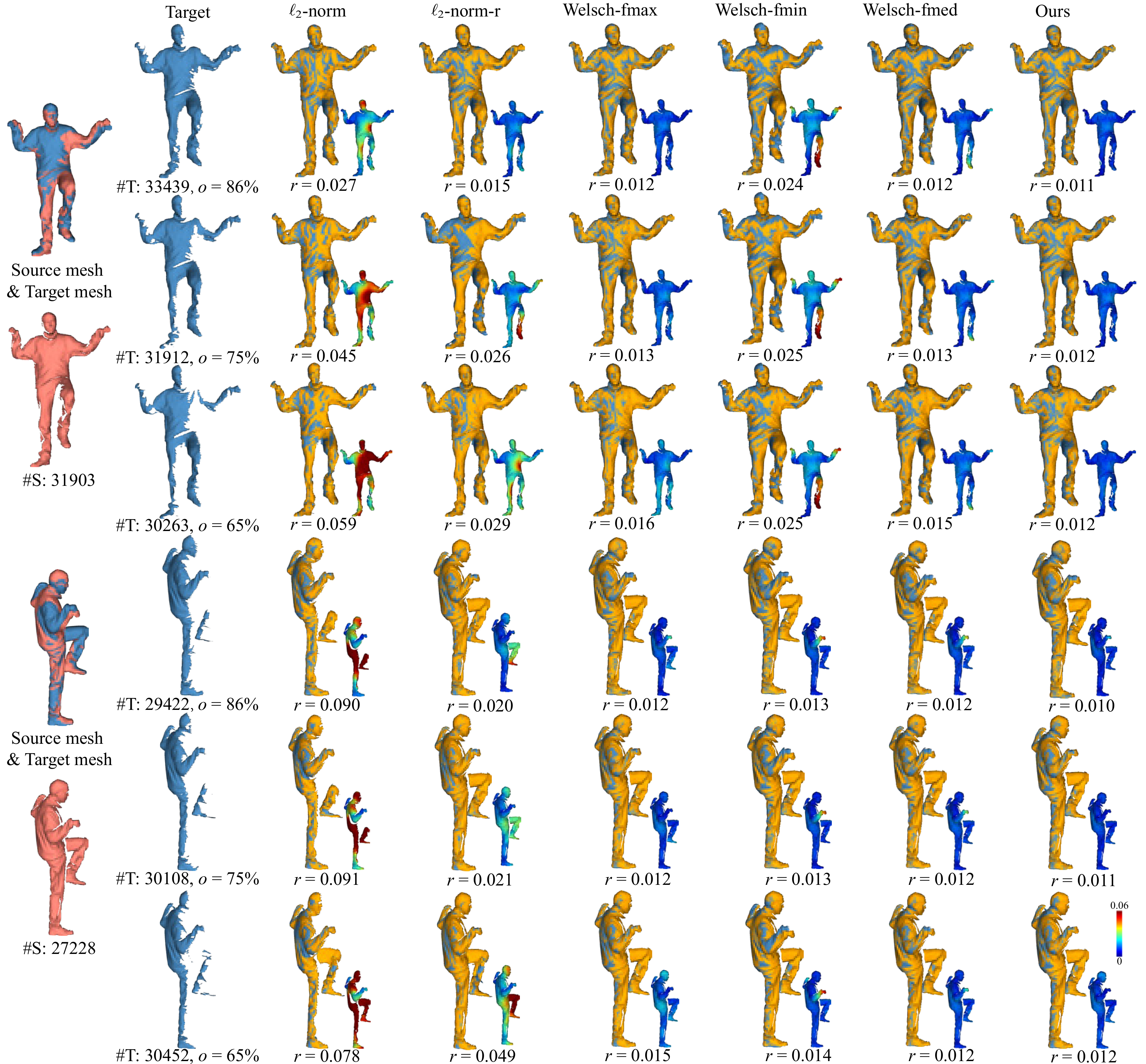}
	\caption{Comparison on partially overlapping data constructed from the ``crane'' dataset~\cite{vlasic2008articulated}. We set $k_{\alpha}=100, k_{\beta}=10$ for all variants.}
	\label{Fig:welsch_effect}
\end{figure*}

We conduct extensive experiments to evaluate the effectiveness of our method and compare its performance with existing approaches.

\subsection{Implementation and metrics}
We implement our method in C++, using Eigen~\cite{eigenweb} for linear algebra operations and OpenMP for parallelization. Unless stated otherwise, the experiments are run on a PC with 16GB of RAM and a 6-code CPU at 3.60GHz. Each pair of surfaces are pre-processed by scaling them to achieve a unit-length diagonal for their bounding box. By default we set $\epsilon=10^{-5}, I_{\max}=100$, and the sampling radius $R=5\overline{l}$ where $\overline{l}$ is the average edge length of the source model. 
The source code of our method is available at \url{https://github.com/yaoyx689/AMM_NRR}.

\begin{figure*}[t]
	\centering
	\includegraphics[width=\textwidth]{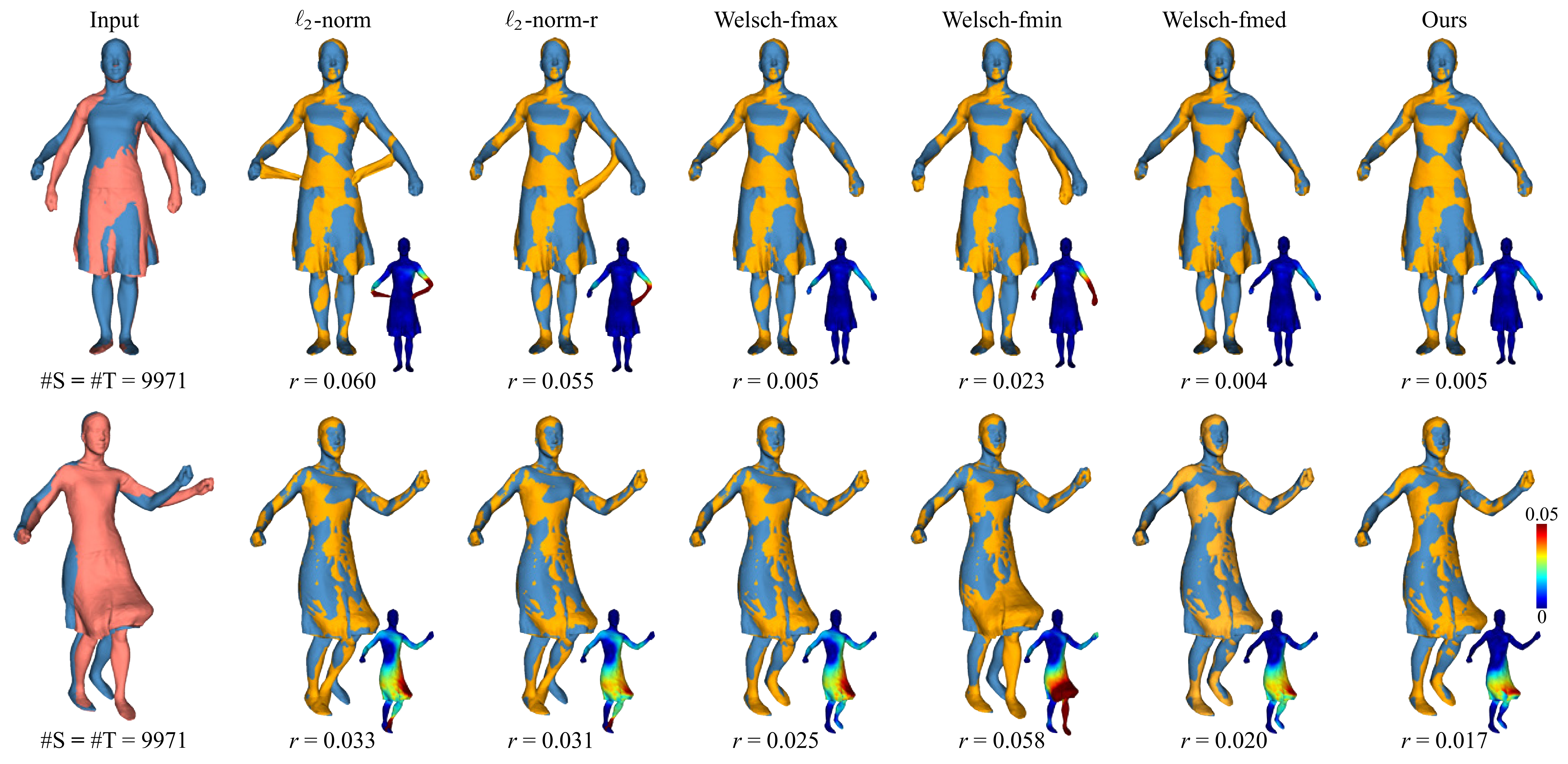}
	\caption{Comparison on two pairs of meshes from the ``samba'' dataset~\cite{vlasic2008articulated}. We set $k_{\alpha}=10, k_{\beta}=1$ for all variants.}
	\label{Fig:welsch_effect2}
\end{figure*}

To evaluate the registration accuracy of a result, we first compute the error for each source point $\mathbf{v}_i$ using the distance from its transformed position $\newpos{\mathbf{v}}_i$ to its ground-truth corresponding point $\mathbf{u}_{\rho^*(i)}$ on the target surface: 
\begin{equation}
    D_i = \|\newpos{\mathbf{v}}_i - \mathbf{u}_{\rho^*(i)}\|.
    \label{eq:PointError}
\end{equation}
The registration accuracy is then measured by the root mean square error (RMSE):
\begin{equation}
r=\sqrt{\frac{\sum_{\mathbf{v}_i \in \mathcal{V}}D_i^2}{|\mathcal{V}|}},
\label{Eq:gt_metric}
\end{equation}
where a lower value of $r$ indicates higher accuracy.  
The error metrics $D_i$ and $r$ are measured in the original scale of the models, and are in  the unit of meters unless stated otherwise.
In the following, we use \#S and \#T to denote the number of sample points on the source surface and the target surface, respectively. 
We render the initial source model, the target model, and the deformed source model in red, blue, and orange respectively, and use color-coding to visualize the point-wise error $D_i$ across the surface.

\subsection{Effectiveness of our method}

\para{Welsch's function and parameters update} We first evaluate the effectiveness of our formulation with Welsch's function and our dynamic update strategies for its parameter. To do so, we consider the following variants of our algorithm:
\begin{itemize}[leftmargin=*]
    \item We replace the Welsch's functions in the target function in Eq.~\eqref{Eq:objective_func} with the squared $\ell_2$-norm, such that the target function becomes
    \[
E_{\ell_2}=\sum_{\mathbf{v}_i\in{\mathcal{V}}}w_i^d\|\newpos{\mathbf{v}}_i-\mathbf{u}_{\rho(i)}\|^2 
+\alpha\sum_{\mathbf{p}_i\in\mathcal{V}_{\mathcal{G}}}\sum_{\mathbf{p}_j\in\mathcal{N}(\mathbf{p}_i)}\|\mathbf{D}_{ij}\|^2 + \beta E_{\text{rot}},
\]
where $\{w_i^d\}$ are weights. We consider two variants with different weighting schemes:
\begin{itemize}
\item All weights $\{w_i^d\}$ are set to 1. In the following examples, we refer to this variant as \emph{$\ell_2$-norm}.
\item The weight $w_i^d$ is set to 1 only if $\newpos{\mathbf{v}}_i$ is close enough to $\mathbf{u}_{\rho(i)}$. Otherwise, it is set to zero. To mimic the exclusion mechanism of our Welsch formulation, we use the empirical three-sigma rule for the Gaussian function and set $w_i^d  = 0$ if  $\|\newpos{\mathbf{v}}_i-\mathbf{u}_{\rho(i)}\| \geq \nu_a$ where $\nu_a$ is the Welsch's function parameter used in our alignment term, i.e., the weight ${w_i^d}^{(k)}$ at the $k$-th iteration is chosen as
\[
{w_i^d}^{(k)} = 
\left\{
\begin{aligned}
1 & \qquad\|\newpos{\mathbf{v}}_i-\mathbf{u}_{\rho(i)}^{(k)}\| < 3\nu_a, \\
0 & \qquad\text{others},
\end{aligned}
\right.
\]
We refer to this variant as \emph{$\ell_2$-norm-r}.
\end{itemize}

\item Instead of dynamically updating $\nu_a$ and $\nu_r$, we fix them throughout the optimization. We consider two variants:
\begin{itemize}
    \item Both parameters are fixed to their maximum values during our dynamic update scheme.  We refer to this variant as \emph{Welsch-fmax}.
    \item Both parameters are fixed to their minimum values during our dynamic update scheme. We refer to this as \emph{Welsch-fmin}.
    \item Both parameters are fixed to their average values between the minimum values and the maximum values during our dynamic update scheme. We refer to this as \emph{Welsch-fmed}.
\end{itemize}
\end{itemize}

In Fig.~\ref{Fig:welsch_effect}, we compare different variants on problems where the source and target models overlap partially. To this end, we choose two pairs of meshes from the ``crane'' dataset of~\cite{vlasic2008articulated}, each consisting of two consecutive frames from the sequence. 
For each pair, we choose a view direction and one of the meshes, and use the visible part of the mesh from the view direction as the source model. For the other mesh, we choose three view directions different from the source model, and use its visible parts from these directions to create three target models that overlap partially with the source model.
Fig.~\ref{Fig:welsch_effect} presents the registration results from different variants on these models.
To compute the RMSE (Eq.~\eqref{Eq:gt_metric}) in this scenario, we note that the point correspondence between the complete models is known from the dataset; thus for each point on the source model we find its corresponding point on the complete mesh of the target model as the ground-truth correspondence for the RMSE evaluation.
For each target model, we also evaluate its overlap ratio $o$ with the source model as the percentage of source points whose ground-truth correspondence lies on the target model. 
Fig.~\ref{Fig:welsch_effect} shows that the $\ell_2$-based variants result in lower accuracy than other variants, due to the sensitivity of $\ell_2$-norm to large errors, while the Welsch-based variants with fixed $\nu_a$ and $\nu_r$ are less effective in distinguishing non-overlapping points. 
Our Welsch-based formulation with a dynamic update of $(\nu_a,\nu_r)$ achieves better accuracy than these variants.

\begin{figure}[bt]
	\centering
	\includegraphics[width=\columnwidth]{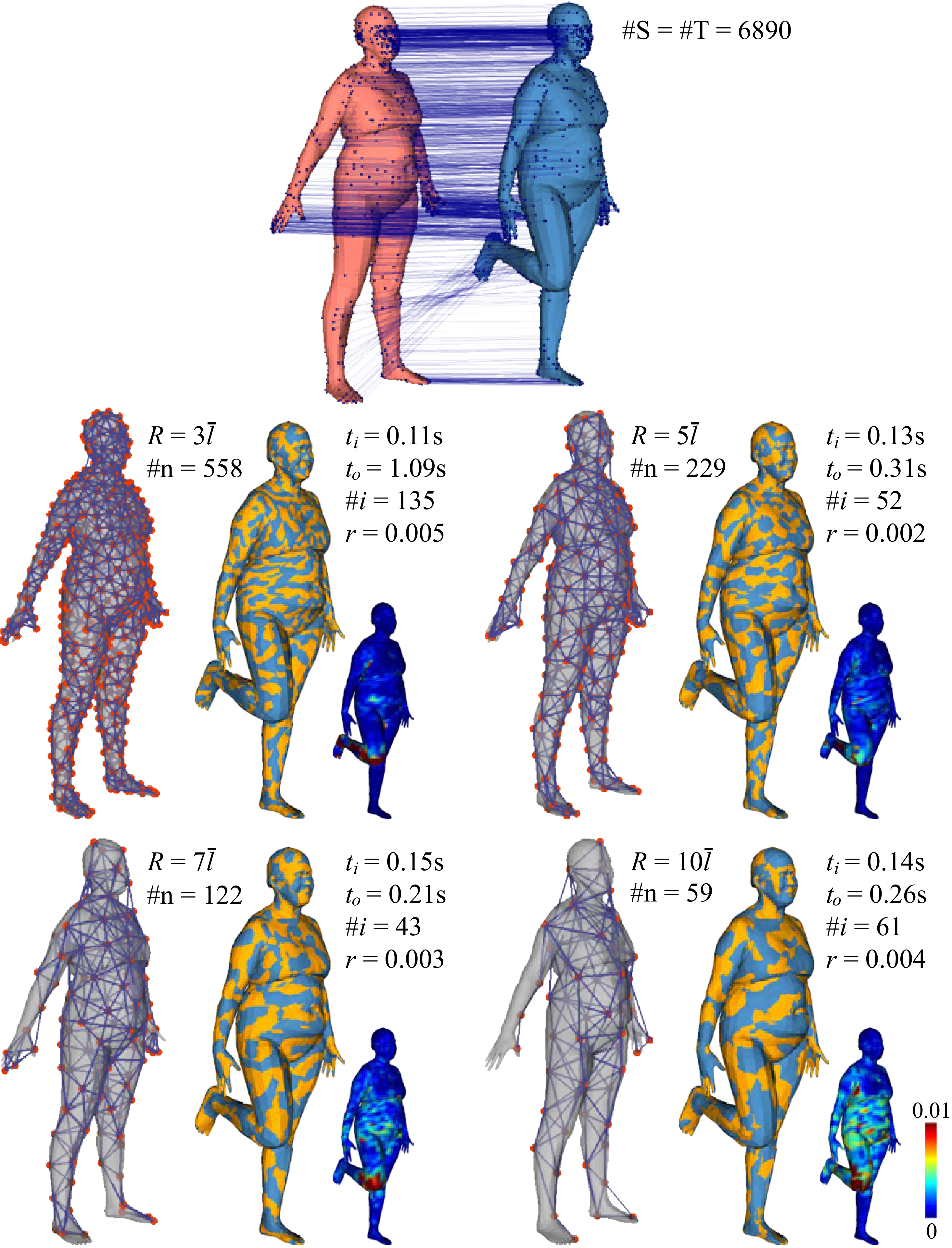}
	\caption{Deformation graphs and registration results using our method with different values of the radius parameter $R$, on a pair of models from the MPI Faust dataset~\cite{bogo2014faust}. We set $k_{\alpha}=0.1,k_{\beta}=0.001$. Here $\#\text{n}$ denotes the number of the deformation graph nodes, $t_i$ and $t_o$ denote the computational time for the deformation graph construction and the numerical optimization respectively, and $\#i$ denotes the number of iterations required for the numerical optimization solver to converge.}
	\label{fig:diff_radius}
\end{figure}

In Fig.~\ref{Fig:welsch_effect2}, we further compare the variants on two pairs of meshes from the ``samba'' dataset~\cite{vlasic2008articulated} with a relatively large deformation between the source and the target.
Our approach also achieves better accuracy than the variants in this scenario, thanks to its use of Welsch's function which helps to attenuate the influence of incorrect correspondence obtained from the closest point, as well as its dynamic parameter update strategy that helps to avoid undesirable local minima of the optimization.

\para{Radius parameter for deformation graph}
The radius parameter ${R}$ for the deformation graph construction affects the number of graph nodes. It can be used to trade off the degrees of freedom and computational efficiency.
Fig.~\ref{fig:diff_radius} shows results using different values of  ${R}$ on a pair of models from the MPI Faust dataset~\cite{bogo2014faust}. We can see that smaller ${R}$ leads to more nodes in the deformation graph, which provides more degrees of freedom for the deformation and allow for more accurate alignment. On the other hand, a larger number of nodes increase the number of variables for the optimization, which requires a longer computational time for the deformation graph construction and the numerical solver.
In addition, having too many variables may cause the optimization to converge to a suboptimal solution, as it becomes more difficult to maintain the local shape (e.g., see the right leg of the result from ${R} = 3 \overline{l}$).
In our experiments, our default choice ${R} = 5 \overline{l}$ usually achieves a good balance between accuracy and efficiency.

\subsection{Comparison with the solver in~\cite{yao2020quasi}}
\label{sec:LBFGSComparison}
\begin{figure}[t]
	\centering
	\includegraphics[width=\columnwidth]{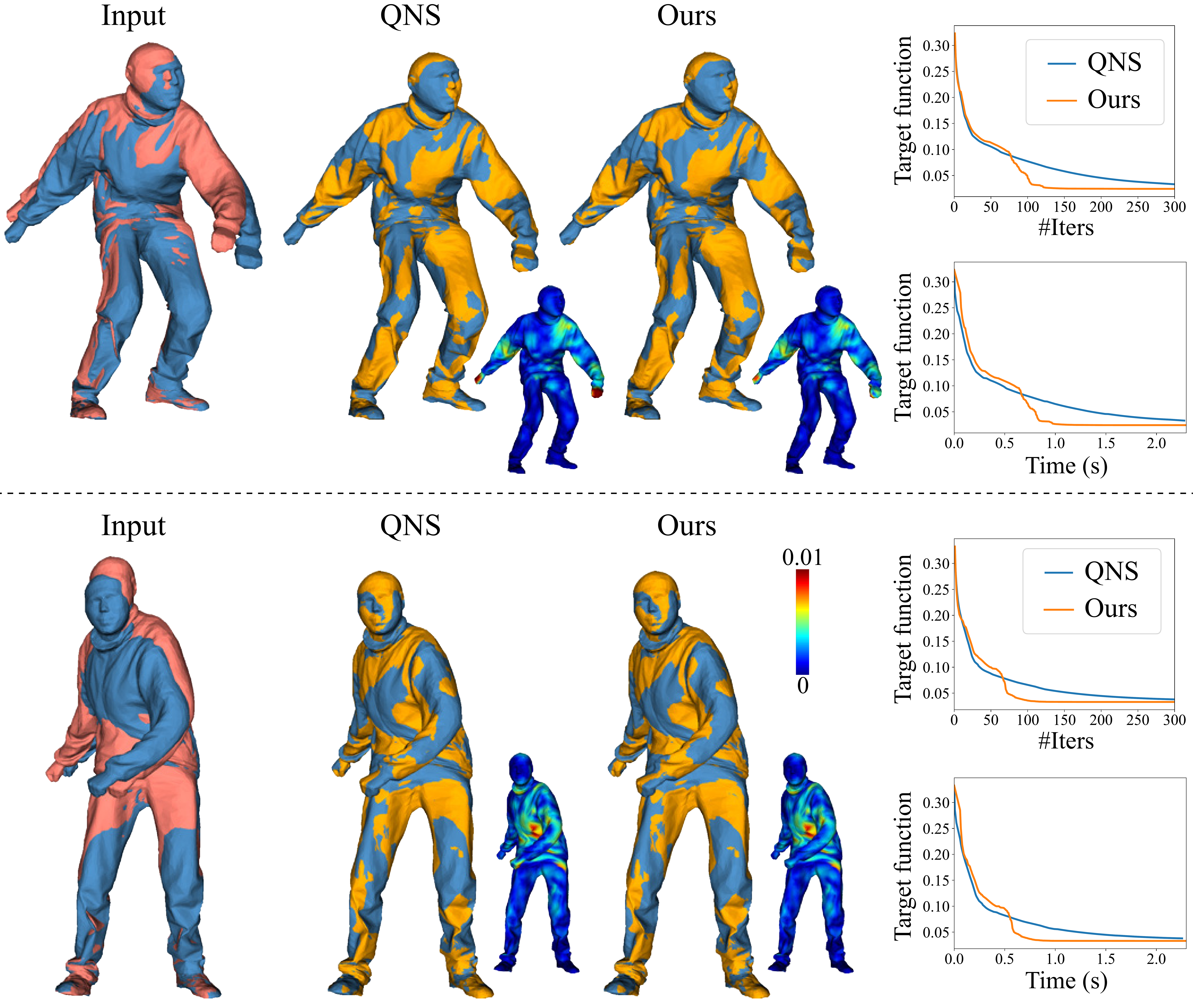}
	\caption{Comparisons between the solver from~\cite{yao2020quasi} and our solver for the same optimization problem on the ``jumping'' dataset of~\cite{vlasic2008articulated}. We set $k_{\alpha}=10,k_{\beta}=100$ for both solvers.}
	\label{fig:compare_aa}
\end{figure}

A preliminary version of our method was presented in~\cite{yao2020quasi}, where the same optimization problem is solved using a quasi-Newton approach instead. Similar to our solver, the solver in~\cite{yao2020quasi} constructs a surrogate for the target function and minimizes it to update the variables, with a dynamic update of the parameters $(\nu_a, \nu_r)$. However, their surrogate function is not quadratic and requires an inner L-BFGS solver for its minimization. 
The two solvers achieve similar results but differ in their efficiency, as shown in Fig.~\ref{fig:compare_aa}.
Here we choose two pairs of adjacent frame data from the ``jumping'' dataset of~\cite{vlasic2008articulated}, and perform registration with fixed parameters $(\nu_a, \nu_r)$ for a fair comparison between the solvers for their speed of convergence.
The plots of the target function show that our solver converges faster than~\cite{yao2020quasi}.
This is partly because the surrogate function is only tight around the current variable values; the L-BFGS inner solver in~\cite{yao2020quasi} spends multiple iterations to minimize a loose upper bound function away from the current variable values, which is less efficient in reducing the target function.
In comparison, our solver immediately updates the surrogate after each iteration to maintain a tight upper bound function for minimization, which improves efficiency. 
Moreover, the use of Anderson acceleration helps our solver to converge faster to the fixed point and avoid the slow decrease of target function as seen in the final phase for the solver of~\cite{yao2020quasi}.

\begin{figure}[t]
	\centering
	\includegraphics[width=\columnwidth]{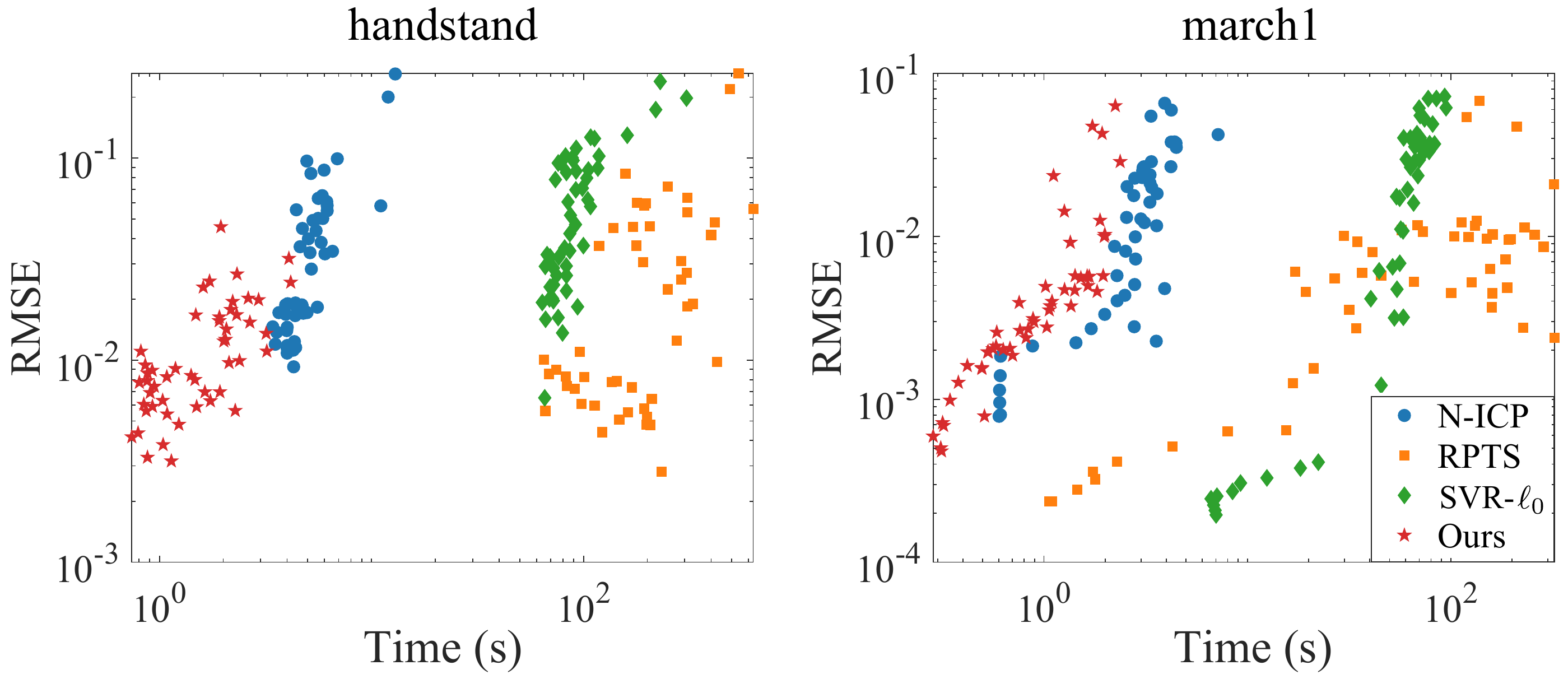}
	\caption{Visualization of the computational time and RMSE using different methods on 50 pairs of models from the ``handstand'' and ``march1'' datasets~\cite{vlasic2008articulated}.}  
	\label{fig:clean_time_rmse}
\end{figure}

\begin{figure}[t]
	\centering
	\includegraphics[width=\columnwidth]{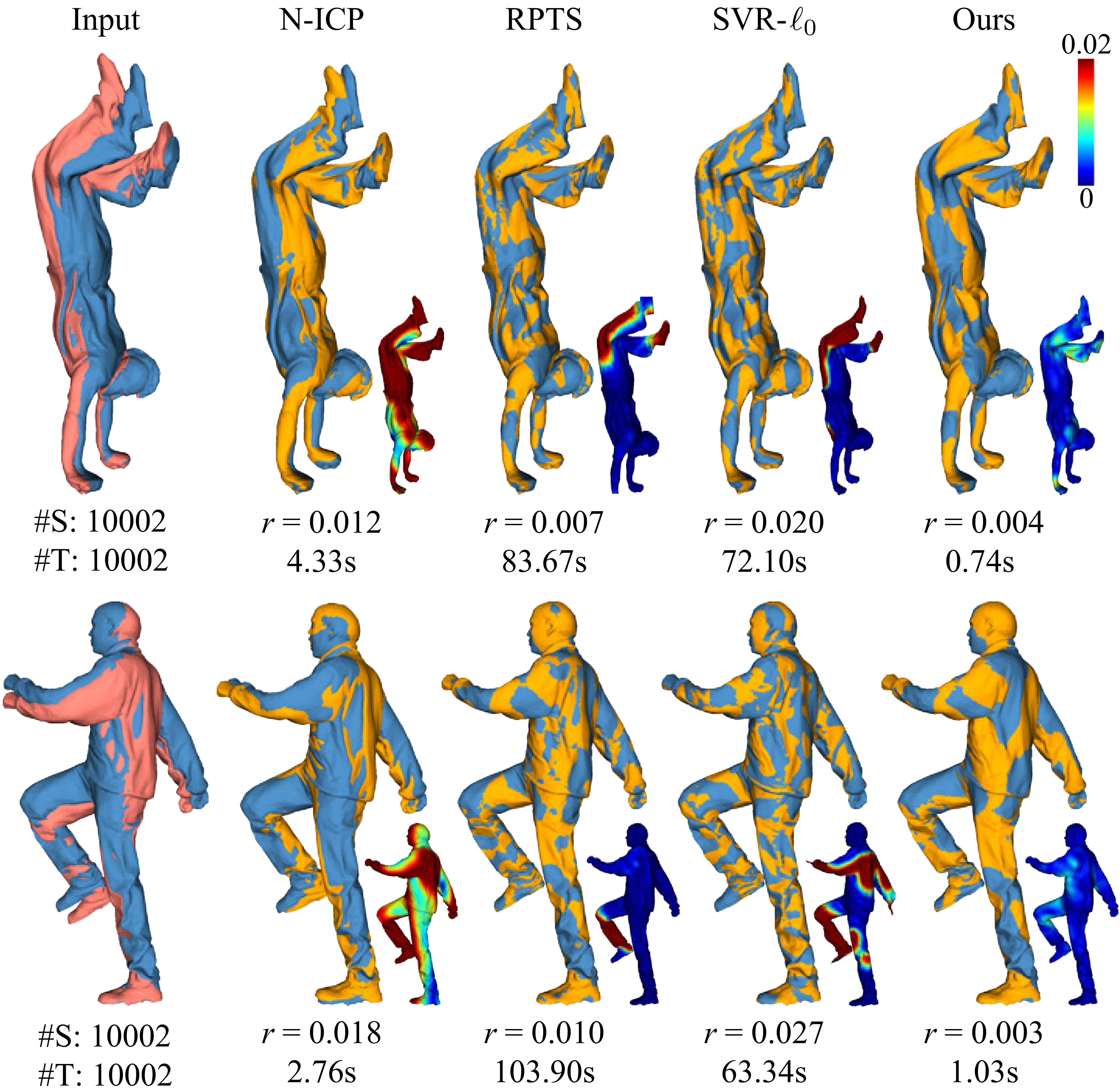}
	\caption{Comparison on mesh pairs from the ``handstand'' dataset (top) and the ``march1'' dataset (bottom)~\cite{vlasic2008articulated}. We set $\alpha=10$ for N-ICP, $\alpha=10, \beta = 1$ for RPTS, $\alpha=1, \beta=1$ for SVR-$\ell_0$, and $k_{\alpha}=100, k_{\beta} = 1$ for our method.}
	\label{Fig:clean_data1}
\end{figure}

\subsection{Comparison with existing methods}

We compare our method with the $\ell_2$-based approach from~\cite{amberg2007optimal} (N-ICP), the $\ell_1$-based approach from~\cite{li2018robust} (RPTS), and the $\ell_0$-based approach from~\cite{guo2015robust} (SVR-$\ell_0$). We implement N-ICP and RPTS by ourselves\footnote{\url{https://github.com/Juyong/Fast_RNRR}}, while the implementation of SVR-$\ell_0$ is provided by the authors.
All these methods solve an optimization problem for the transformation, where the target function is a weighted sum of an alignment term, a regularization term, and (for some methods) a local rotation/rigidity term.
To unify notation, for each method we set the weight for the alignment term to 1 unless stated otherwise, and use $\alpha$ and $\beta$ to denote the weight for the regularization term and the rotation/rigidity term, respectively.
For a fair comparison, we choose the best parameters for each method.
For our method, to reduce sensitivity to the mesh resolution and the Welsch's function parameters $\nu_a, \nu_r$, we set the weights $\alpha, \beta$ as
\begin{equation}
    \alpha=k_{\alpha}\cdot \frac{|\mathcal{V}|}{|\mathcal{E}_{\mathcal{G}}|}\cdot\frac{\nu_r^2}{\nu_a^2}, \quad 
    \beta=k_{\beta} \cdot \frac{|\mathcal{V}|}{|\mathcal{V}_{\mathcal{G}}|} \cdot\frac{1}{2\nu_a^2},
    \label{eq:WeightSettings}
\end{equation}
and control $\alpha, \beta$ using the parameters $k_{\alpha}$ and $k_{\beta}$.
For N-ICP, we ignore the parameter adjustment for the stiffness and landmark terms for a fair comparison. 
For SVR-$\ell_0$, we extract the $\ell_0$-optimization part and replace the corresponding points with the closest points.
For both N-ICP and RPTS, we terminate the solver if it reaches 100 iterations or  $\max_i\|\hat{\mathbf{v}}_i^{(k+1)}-\hat{\mathbf{v}}_i^{(k)}\|<1 \times 10^{-3}$.
RPTS also requires an inner ADMM solver to compute all affine transformations $\mathbf{X}$, and it is terminated when reaching 20 iterations or $\|\mathbf{X}^{(k+1)} - \mathbf{X}^{(k)}\|_F < 1 \times 10^{-3}$.
For SVR-$\ell_0$, we terminate the solver when it reaches 100 iterations or  $\|\mathbf{X}^{(k+1)} - \mathbf{X}^{(k)}\|_F < 1 \times 10^{-3}$.
We compare the methods on the human motion datasets from~\cite{vlasic2008articulated}, the MPI Faust dataset~\cite{bogo2014faust}, the TOSCA dataset~\cite{bronstein2008numerical}, the DeepDeform dataset~\cite{bozic2020deepdeform}, and real-world data acquired by a Kinect camera.
The results are presented in detail in the following paragraphs. The iterative registration process of different optimization-based methods on some problem instances is also demonstrated in the supplementary video\footnote{\url{https://youtu.be/1v3ggyoUxT0}}.

\para{Clean data with small deformation}
In Figs.~\ref{fig:clean_time_rmse} and \ref{Fig:clean_data1} and Tab.~\ref{Tab:compare_clean},
we test the methods on the ``handstand'' and ``march1'' datasets from~\cite{vlasic2008articulated}.
Each dataset contains a sequence of meshes for the motion of a human performer. We register the $i$-th frame to the ($i$+$2$)-th frame for each dataset. In this setting, the deformation between the source and the target models is relatively small. 
Fig.~\ref{fig:clean_time_rmse} shows the computational time and RMSEs (Eq.~\eqref{Eq:gt_metric}) of 50 pairs of such models, while Fig.~\ref{Fig:clean_data1} shows the results on one example pair for each dataset. 
Tab.~\ref{Tab:compare_clean} further shows the mean and median values of RMSE for each dataset.
The results show that overall our method can achieve higher accuracy with a shorter computational time.

\begin{table}[t]
	\caption{Mean/median RMSE ($\times 10^{-2}$) and average computational time (s) using different methods on 50 pairs of models from the ``handstand'' and ``march1'' datasets~\cite{vlasic2008articulated}. We set $\alpha=10$ for N-ICP, $\alpha=100, \beta=1$ for RPTS, $\alpha=1, \beta=1$ for SVR-$\ell_0$, and $k_{\alpha}=100, k_{\beta}=1$ for our method.}
	\label{Tab:compare_clean}
		\setlength{\tabcolsep}{3.8pt}
	\centering
	\begin{small}
		\begin{tabular}{ c  c  c  c  c  c}
			\Xhline{1pt}
			\multicolumn{1}{c}{Data} & & N-ICP
			&RPTS & SVR-$\ell_0$ 		
			& Ours  \\\hline
			\multirow{3}{*}{handstand} & Mean RMSE  &4.31	&3.30	&6.45		&\textbf{1.21}	\\
			& Median RMSE	&3.09	&1.17	&4.95		&\textbf{0.90}	\\
			&Average Time	&5.31	&215.47	&96.61		&\textbf{1.70}	\\\hline
			\multirow{3}{*}{march1} & Mean RMSE	&1.78	&0.93	&2.49		&\textbf{0.75}	\\
			&Median RMSE	&1.46	&0.60	&2.51		&\textbf{0.33}	\\
			&Average Time 	&2.86	&103.12	&55.36	&\textbf{1.06}	\\
			\Xhline{1pt}
		\end{tabular}
	\end{small}
\end{table}

\begin{figure}[!t]
	\centering
	\includegraphics[width=\columnwidth]{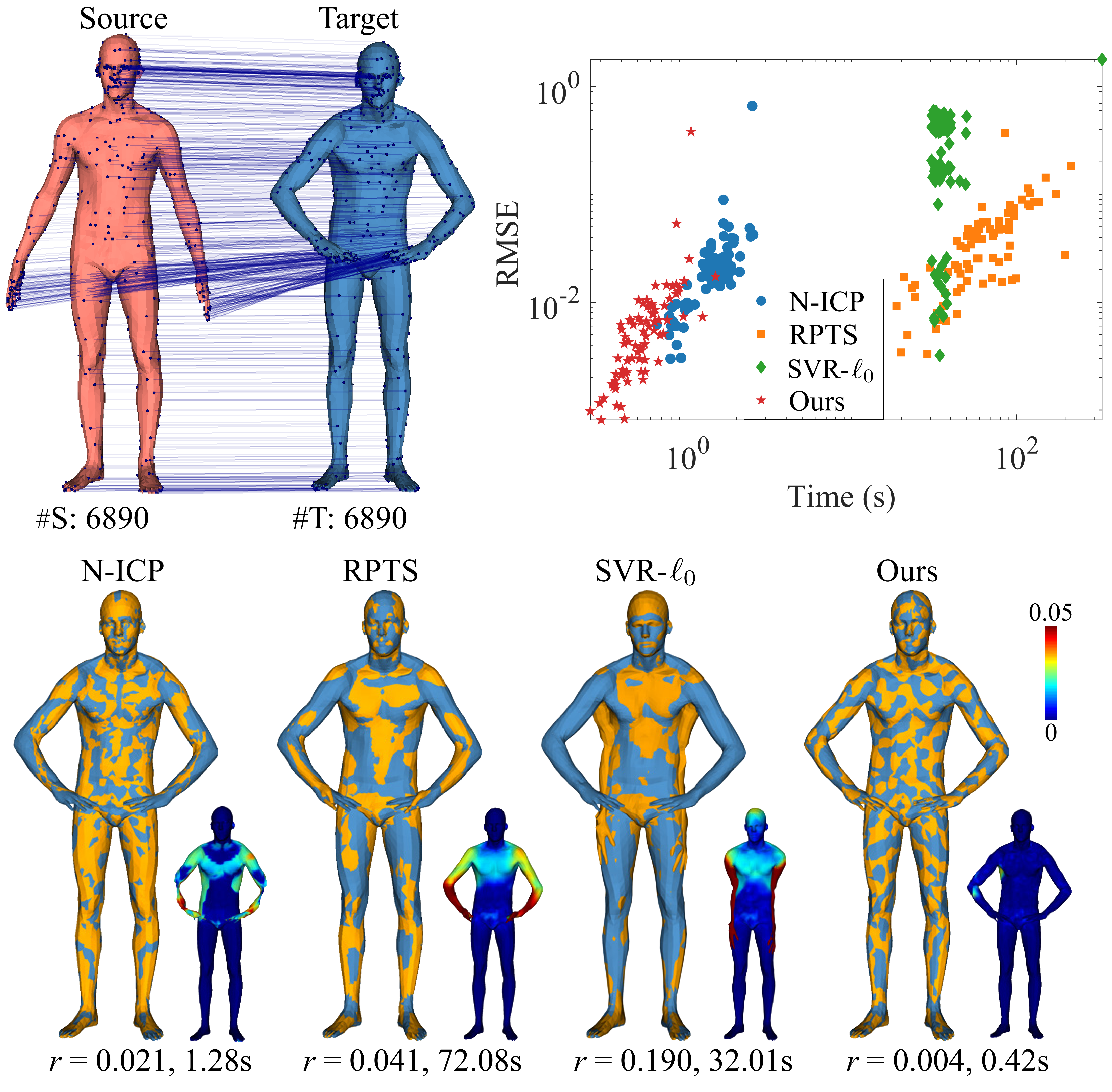}
	\caption{Comparisons on the MPI Faust dataset~\cite{bogo2014faust}. The plot in the upper right visualizes the computational time and RMSE using different methods on 80 pairs of models from the dataset, while the rendered images show the results on a particular pair. }
	\label{Fig:faust_time_rmse}
\end{figure}

\begin{table}[t]
	\caption{Mean RMSE ($\times 10^{-3}$) and average computational time (s) using different methods on models from the MPI Faust dataset~\cite{bogo2014faust}.  We set $\alpha=0.1$ for N-ICP, $\alpha=0.1, \beta=0.001$ for RPTS, 
	$\alpha=0.001, \beta=1$ for SVR-$\ell_0$, 
	and $k_{\alpha}=0.1, k_{\beta}=0.001$ for our method.}
	\label{Tab:big_deform}
	\setlength{\tabcolsep}{1.6pt}
	\centering
	\begin{small}
		\begin{tabular}{ c  c  c  c  c  c  c c c}
			\Xhline{1pt}
			\multirow{2}{*}{Pose pair} & \multicolumn{2}{c}{N-ICP} & 
			\multicolumn{2}{c}{RPTS} 
			&\multicolumn{2}{c}{SVR-$\ell_0$} 
			& \multicolumn{2}{c}{Ours} \\\cline{2-9}
			& Time
			& \makecell[c]{RMSE}	& Time
			& \makecell[c]{RMSE }	& Time
			& \makecell[c]{RMSE }	& Time
			& \makecell[c]{RMSE}	\\\hline
			1	&1.47	&21.30	&67.18	&43.77 &35.84	&170.75	&\textbf{0.52}	&\textbf{4.74}	\\
			2	&0.97	&9.69	&35.52	&14.54 &35.13	&34.08	&\textbf{0.40}	&\textbf{2.21}	\\
			3	&1.44	&26.29	&88.58	&65.32 &36.60	&406.23	&\textbf{0.65}	&\textbf{7.36}	\\
			4	&1.85	&20.69	&86.69	&23.83 &35.02	&166.94	&\textbf{0.59}	&\textbf{4.50}	\\
			5	&0.84	&5.60	&29.64	&6.96 &34.31	&52.13	&\textbf{0.34}	&\textbf{1.25}	\\
			6	&1.59	&19.85	&56.75	&23.68 &63.55	&541.82	&\textbf{0.61}	&\textbf{7.22}	\\
			7	&2.05	&110.15	&125.90	&132.49 &34.84	&487.12	&\textbf{0.96}	&\textbf{55.75}	\\
			8	&1.53	&22.87	&66.29	&47.80	&36.25	&542.07 &\textbf{0.57}	&\textbf{8.46}	\\\hline
			Mean & 
			1.47 & 29.56 & 69.57 & 44.80 & 38.94 & 300.14 & \textbf{0.58} & \textbf{11.44}\\
			Median & 1.50 & 21.00 & 66.74 & 33.80 & 35.49 & 288.49 & \textbf{0.58} & \textbf{5.98} \\	
			\Xhline{1pt}
		\end{tabular}
	\end{small}
\end{table}

\para{Clean data with large deformation} 
In Tab.~\ref{Tab:big_deform} and Fig.~\ref{Fig:faust_time_rmse}, we compare the methods on the MPI Faust dataset~\cite{bogo2014faust}.
We select 10 subjects with nine poses, and use the first pose of each subject as the source surface and the other poses of the same subject as its targets. 
Due to the large difference between the source and the target models, we use the SHOT feature~\cite{tombari2010unique} with diffusion pruning~\cite{tam2014diffusion} to obtain initial corresponding points. 
In Fig.~\ref{Fig:faust_time_rmse}, we visualize the initial correspondences using  blue lines connecting the corresponding points. 
Fig.~\ref{Fig:faust_time_rmse} shows a visualization of the computational time and RMSE for each pose pair using different methods, as well as the results for a particular pair on one subject.
Tab.~\ref{Tab:big_deform} further shows the average RMSE and computational time among all subjects for each pose pair.
We can see that our method requires significantly less computational time while achieving better accuracy. 
A major factor for the efficiency of our method is the adoption of a deformation graph, which only requires optimizing one affine transformation per graph vertex. 
In comparison, N-ICP and RPTS require optimizing one affine transformation
per mesh vertex, which significantly increases the number
of variables as well as computational time.
In addition, the RMSE of SVR-$\ell_0$ is notably higher than the other methods, as it was originally designed for aligning nearby frames for motion tracking and becomes less effective on models with large deformation.

\begin{figure}[t]
	\centering
	\includegraphics[width=\columnwidth]{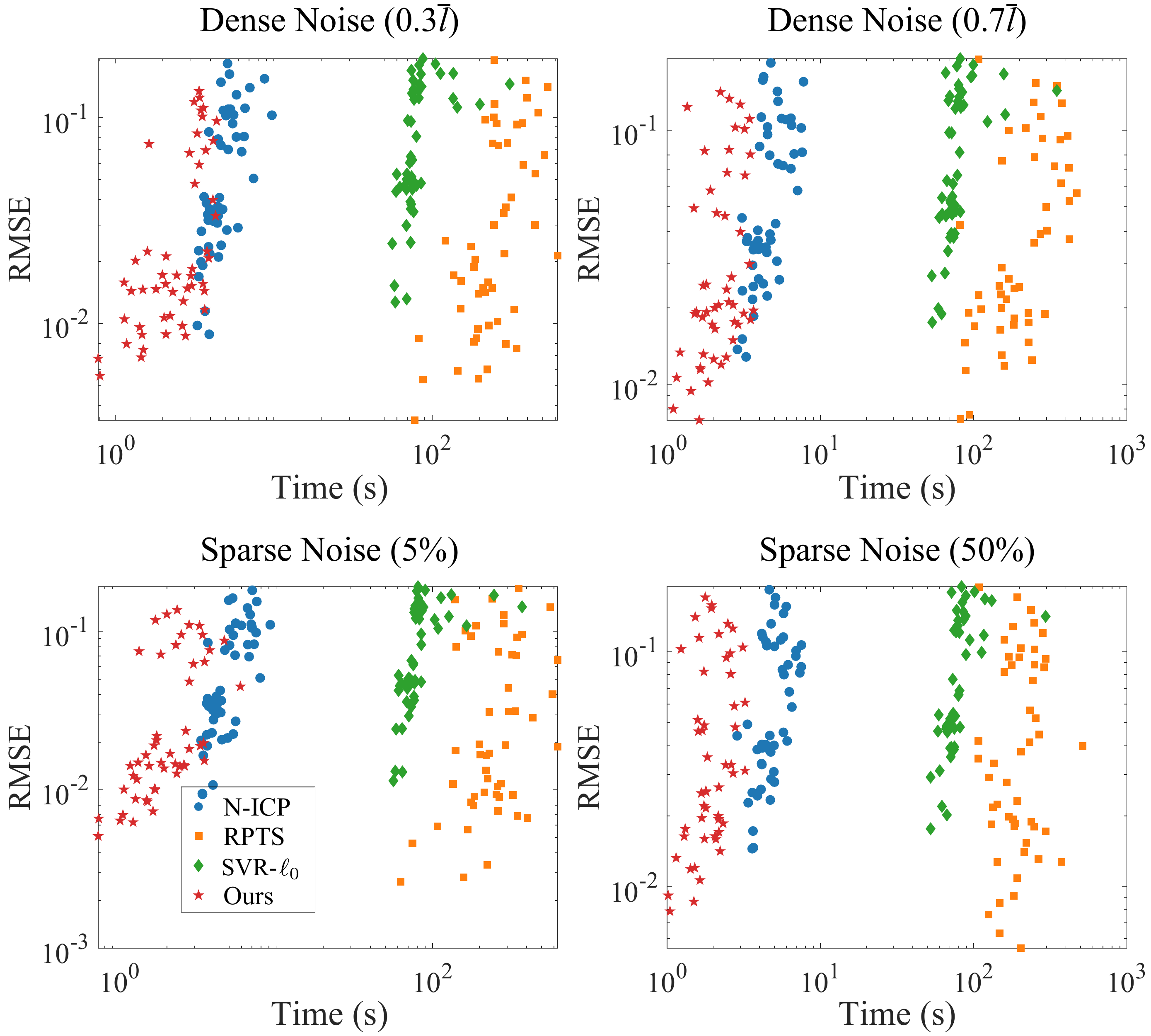}
	\caption{Visualization of computational time and RMSE resulting from different methods on 50 pairs of models constructed from the ``jumping'' dataset~\cite{vlasic2008articulated} with different profiles of added noise.}
	\label{fig:noise_time_rmse}
\end{figure}

\begin{table}[!t]
	\caption{Mean/median RMSE ($\times 10^{-2}$) and average computational time (s) using different methods on 50 pairs of models constructed from the ``jumping'' dataset~\cite{vlasic2008articulated} with different noise profiles. We set $\alpha=10$ for N-ICP,  $\alpha=100, \beta=1$ for RPTS, $\alpha=1, \beta=1$ for SVR-$\ell_0$, and $k_{\alpha}=100, k_{\beta}=1$ for our method.}
	\label{Tab:compare_noise}
	\setlength{\tabcolsep}{3.4pt}
	\centering
	\begin{small}
		\begin{tabular}{ c  c  c  c  c  c}
			\Xhline{1pt}
			\multicolumn{1}{c}{Data} & & N-ICP
			&RPTS & SVR-$\ell_0$ 	
			& Ours  \\\hline
			\multirow{3}{*}{\makecell[c]{Dense~Noise\\(0.3$\overline{l}$)}} & \makecell[c]{Mean RMSE}	&6.39	&4.47	&8.67		&\textbf{3.62}	\\
			& \makecell[c]{Median RMSE}	&3.96	&2.16	&6.13		&\textbf{1.64}	\\
			&Average Time	&4.88	&268.07	&86.19		&\textbf{2.60}\\\hline
			\multirow{3}{*}{\makecell[c]{Dense~Noise\\(0.7$\overline{l}$)}} & \makecell[c]{Mean RMSE  }	&6.62	&5.06	&8.77	&\textbf{4.04}	\\
			& \makecell[c]{Median RMSE }	&4.16	&2.74	&6.25	&\textbf{2.02}	\\
			&Average Time	&4.74	&224.94	&83.50	&\textbf{2.27}	\\\hline
			\multirow{3}{*}{\makecell[c]{Sparse~Noise\\($5\%$)}} & \makecell[c]{Mean RMSE}	&6.41	&4.47	&8.66	&\textbf{3.71} \\
			& \makecell[c]{Median RMSE }	&3.83	&1.82	&6.23	&\textbf{1.59} \\
			&Average Time	&5.09	&270.34	&89.42		&\textbf{2.17}	\\\hline
			\multirow{3}{*}{\makecell[c]{Sparse~Noise\\($50\%$)}} & \makecell[c]{Mean RMSE }	&6.96	&5.37	&8.83	&\textbf{5.23}	\\
			& \makecell[c]{Median RMSE }	&4.47	&3.44	&6.55	&\textbf{3.08}	\\
			&Average Time 	&4.95	&206.29	&82.87	&\textbf{1.99}	\\
			\Xhline{1pt}
		\end{tabular}
	\end{small}
\end{table}

\begin{figure}[!t]
	\centering
	\includegraphics[width=\columnwidth]{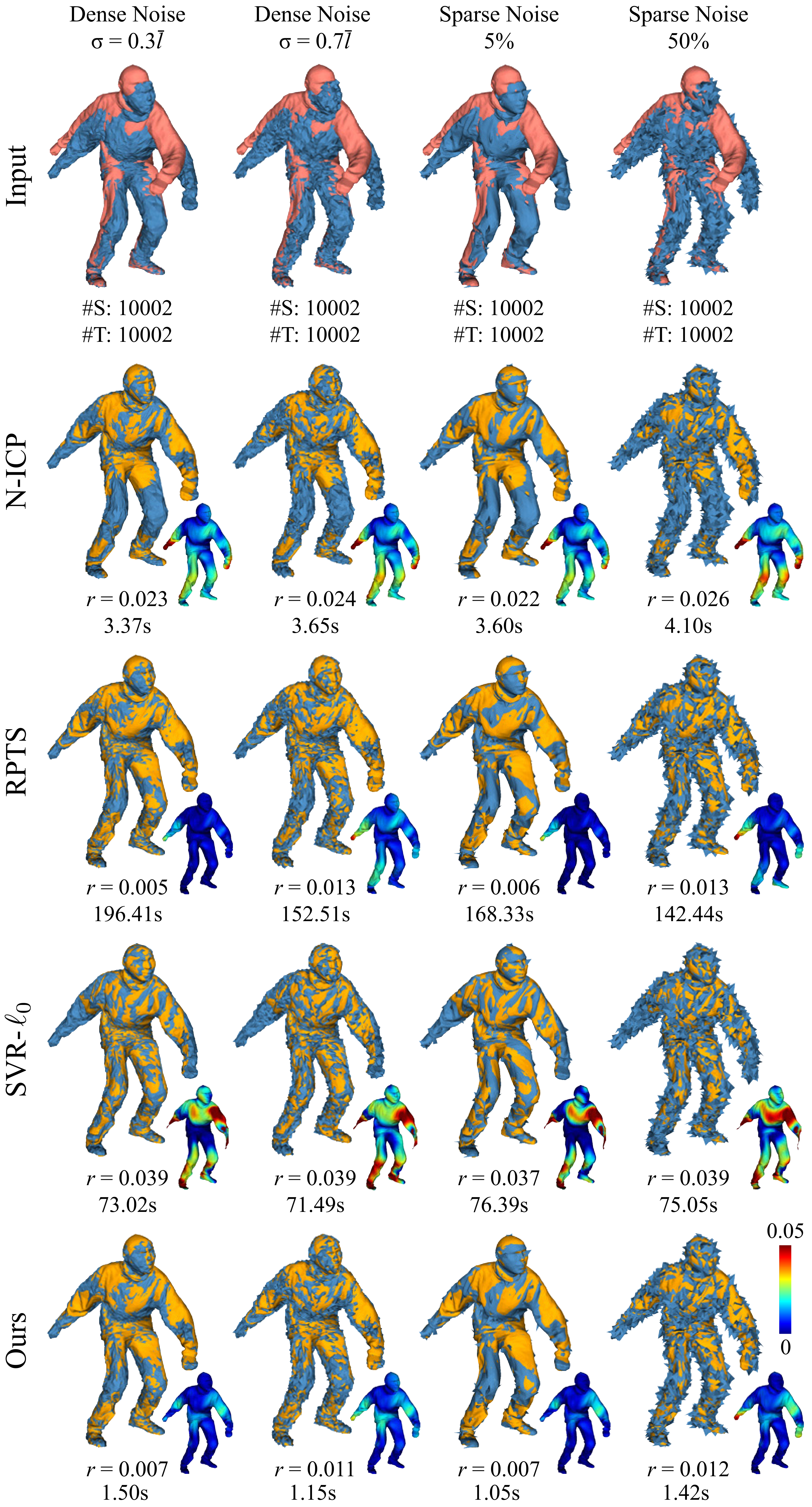}
	\caption{Comparison on a pair of models constructed from the ``jumping'' dataset~\cite{vlasic2008articulated} using different profiles of added noise. We set $\alpha=10$ for N-ICP, $\alpha=100, \beta = 1$ for RPTS, $\alpha=1, \beta = 1$ for SVR-$\ell_0$, and $\alpha=100, \beta = 1$ for our method.}
	\label{Fig:noise}
\end{figure}

\para{Noisy data}
To evaluate the effectiveness of our method on noisy data, we synthesize noisy models using the ``jumping'' dataset from~\cite{vlasic2008articulated}.
Specifically, we collect 50 pairs of models, where each pair consists of the $i$-th frame from the dataset as the source model, and the ($i$+$2$)-th frame with added Gaussian noise as the target model.
For each pair, we generate four noisy variants of the target model as follows:
\begin{itemize}[leftmargin=*]
	\item Dense~Noise~($0.3\overline{l}$): we add noises with a standard deviation $\sigma=0.3\overline{l}$ to all vertices;
	\item Dense~Noise~($0.7\overline{l}$): we add noises with a standard deviation $\sigma=0.7\overline{l}$ to all vertices;
	\item Sparse~Noise~($5\%$): we add noises with a standard deviation $\sigma=\overline{l}$ to $5\%$  of the vertices.
	\item Sparse~Noise~($50\%$): we add noises with a standard deviation $\sigma=\overline{l}$ to $50\%$  of the vertices.
\end{itemize}
Fig.~\ref{fig:noise_time_rmse} visualizes the computational time and RMSE for each method on the 50 pairs of models for each noise profile, while Fig.~\ref{Fig:noise} shows the results on a particular model pair.
Tab.~\ref{Tab:compare_noise} shows for each method the average RMSE and computational time for all model pairs under each noise profile.
We can see that our method achieves better overall performance in terms of both accuracy and efficiency.

\begin{figure}[t]
	\centering
	\includegraphics[width=\columnwidth]{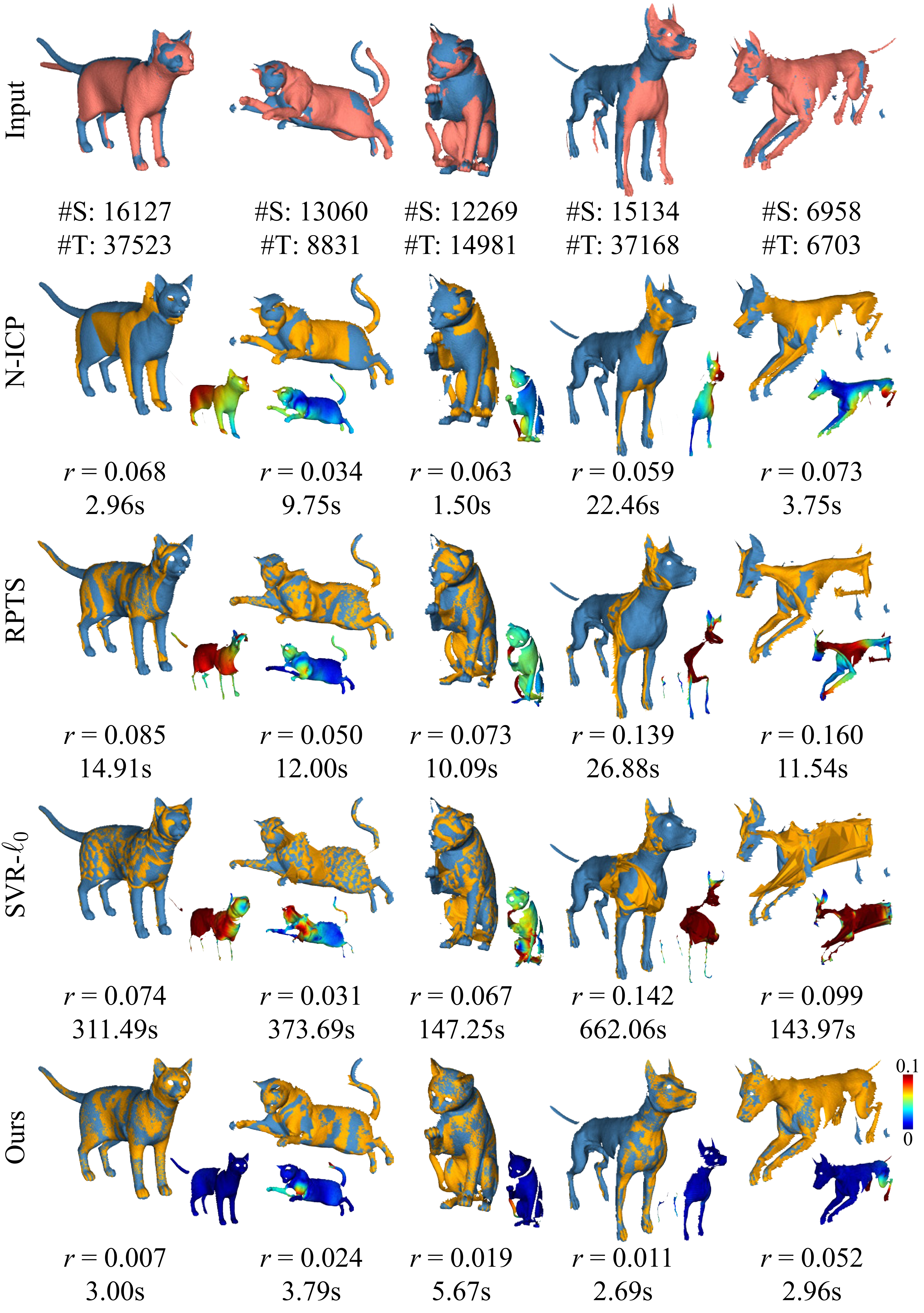}
	\caption{Comparison on partially overlapping data constructed from the TOSCA dataset~\cite{bronstein2008numerical}. We set $\alpha=100$ for N-ICP, $\alpha=1000, \beta = 1$ for RPTS, $\alpha=1, \beta = 1$ for SVR-$\ell_0$, and $\alpha=100, \beta = 10$ for our method. We set $R = 8\overline{l}$ because the mesh is uneven.}
	\label{Fig:TOSCA}
\end{figure}

\begin{figure*}[t]
	\centering
	\includegraphics[width=0.95\textwidth]{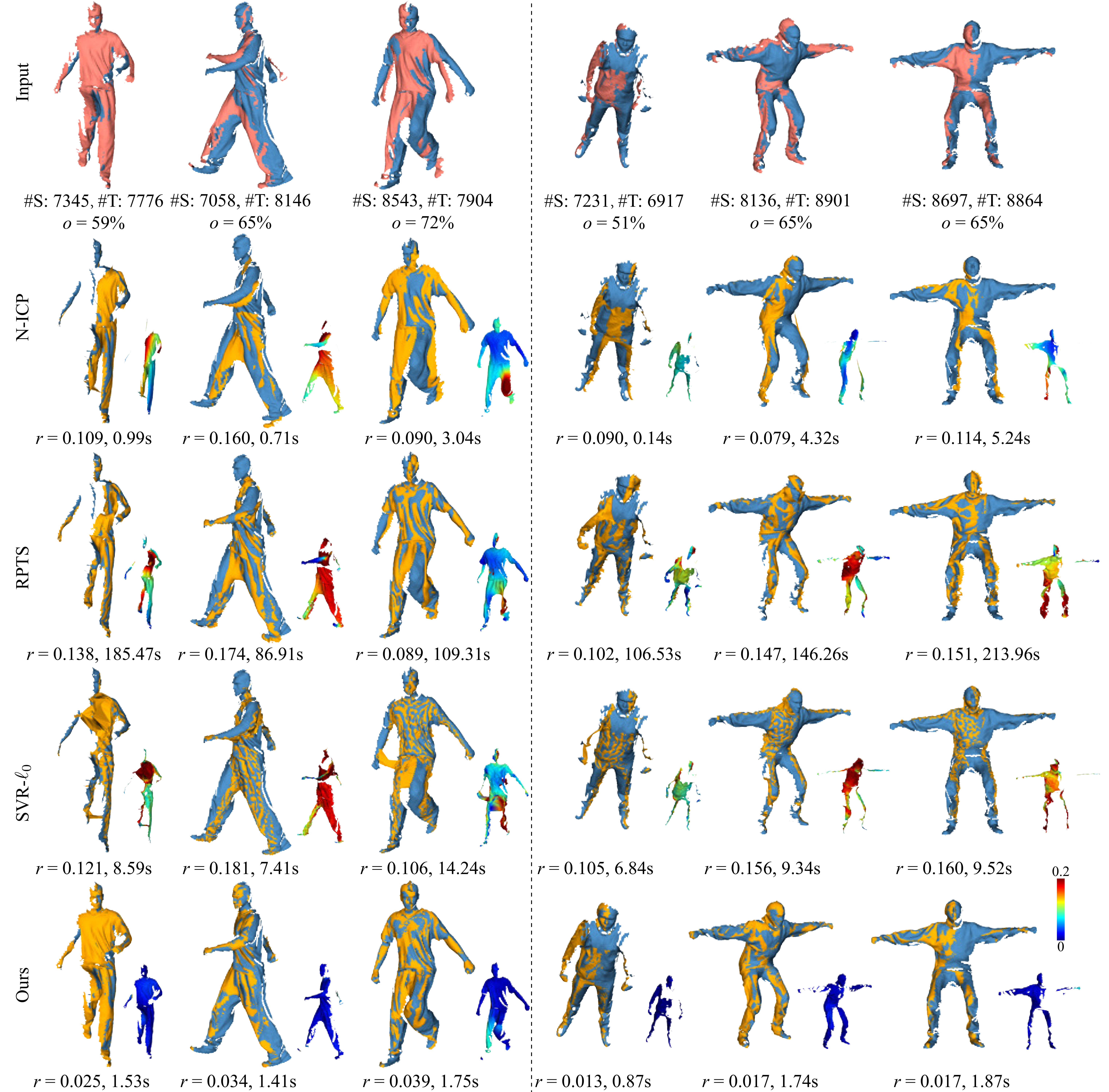}
	\caption{Comparison on six pairs of partially overlapping models constructed from the ``march2'' (left) and ``squat1'' (right) datasets~\cite{vlasic2008articulated}. We set $\alpha=10$ for N-ICP, $\alpha=1000, \beta = 1$ for RPTS, $\alpha=1, \beta = 1$ for SVR-$\ell_0$, and $\alpha=100, \beta = 10$ for our method.}
	\label{Fig:partial}
\end{figure*}

\para{Partial overlaps}
We also evaluate different methods on models that overlap partially.
We choose three pairs of models from the ``march2'' dataset and three pairs of models from the ``squat1'' dataset from~\cite{vlasic2008articulated}, where the models in each pair are two frames apart.
Then for each pair, we construct two different view directions for the two models respectively, and use their visible parts from these directions to construct source and target models that overlap partially when aligned.
We also construct five pairs of models from the TOSCA dataset~\cite{bronstein2008numerical} in a similar way.
Figs.~\ref{Fig:TOSCA} and \ref{Fig:partial} show the results from different methods on these partially overlapping models.
N-ICP performs poorly on such models, since its $\ell_2$-based optimization can be sensitive to target function terms with large residuals. The $\ell_1$-based formulation in RTPS is more robust than an $\ell_2$ formulation, but it still attempts to reduce large residual terms, which can lead to incorrect alignments.
SVR-$\ell_0$ applies an $\ell_0$-based regularization term but its alignment term is still based on the $\ell_2$-norm, which can affect its robustness.
In comparison, our use of the Welsch's function on both the alignment term and the regularization term helps to improve robustness and achieve overall better results in these examples.

\begin{figure*}[t]
	\centering
	\includegraphics[width=\textwidth]{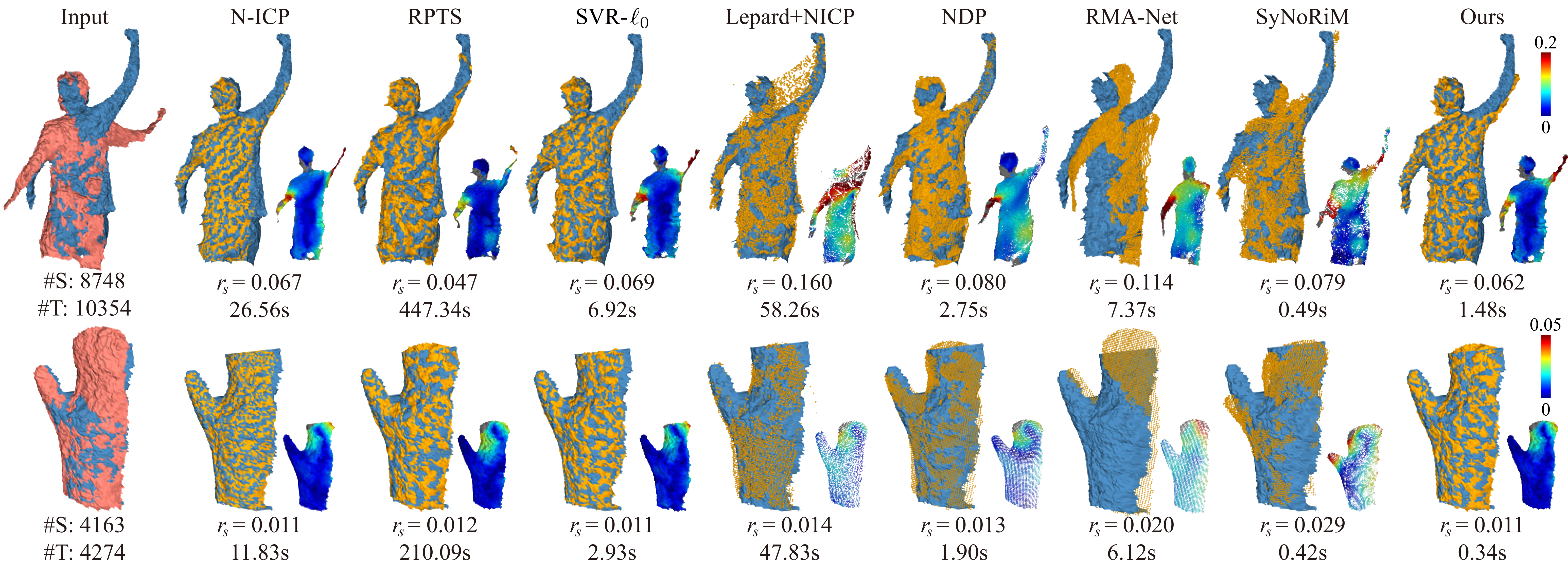}
	\caption{Comparison between different methods on the DeepDeform dataset~\cite{bozic2020deepdeform}. We set $\alpha=10$ for N-ICP, $\alpha=100, \beta=1$ for RPTS, $\alpha=1, \beta=1, R=8\overline{l}$ for SVR-$\ell_0$, and $k_{\alpha}=10, k_{\beta}=1, R=8\overline{l}$ for ours. In the visualized error map, we show the points without ground-truth scene flows in gray.}
	\label{Fig:deepdeform}
\end{figure*}

\begin{table}[t]
	\caption{Mean/median $r_s$ ($\times 10^{-2}$) and average computational time (s) using different methods on 349 pairs of models from DeepDeform dataset~\cite{bozic2020deepdeform}. For Lepard+NICP, there are only 219 valid results.  We set $\alpha=10$ for N-ICP,  $\alpha=100, \beta=1$ for RPTS, $\alpha=1, \beta=1, r=8\overline{l}$ for SVR-$\ell_0$, and $k_{\alpha}=10, k_{\beta}=1, r=8\overline{l}$ for our method.}
	\label{Tab:deepdeform}
	\setlength{\tabcolsep}{3.4pt}
	\centering
	\begin{small}
		\begin{tabular}{ c  c  c  c}
			\Xhline{1pt}
			\multicolumn{1}{c}{Method} & Mean $r_s$ & Median $r_s$ & Average Time \\\hline
N-ICP & 6.58 & 2.00 & 74.61\\
RPTS & 6.49 & 1.88 & 1050.64\\
SVR-$\ell_0$ & 6.67 & 2.07 & 10.47\\
Lepard+NICP & 8.04 & 4.40 & 54.55\\
NDP & 6.66 & \textbf{1.86} & 2.04\\
RMA-Net & 15.72 & 7.63 & 7.56\\
SyNoRiM & 7.90 & 2.75 & \textbf{0.56}\\
Ours & \textbf{6.49} & 1.93 & 1.39\\
			\Xhline{1pt}
		\end{tabular}
	\end{small}
\end{table}

\para{Real scanned data}
Our method is also effective on real-world data.
To demonstrate this, we conduct experiments on the real scanned data from the DeepDeform dataset~\cite{bozic2020deepdeform}. 
This dataset provides RGB-D frame pairs with estimated optical flows and scene flows and manually annotated sparse corresponding point pairs. 
To evaluate the accuracy of the result, we adopt the scene flows provided by the dataset as the ground-truth deformation, and compute an error measure as follows: 
\[
r_{s} = \sqrt{\frac{1}{|\mathcal{S}|}
\sum_{\mathbf{t}_i^s\in \mathcal{S}}\|\hat{\mathbf{v}}_i-(\mathbf{v}_i + \mathbf{t}^{s}_i)\|^2},
\]
where $\mathcal{S}$ is the set of scene flows for the source surface points.
We use 349 pairs in the validation set for testing, as the validation set includes the scene flows required for computing $r_{s}$. 
Besides N-ICP, RPTS and SVR-$\ell_0$, we also compare with four recent deep learning methods, Lepard+NICP~\cite{lepard2021}, NDP~\cite{li2022DeformationPyramid}, RMA-Net~\cite{feng2021recurrent}, and SyNoRiM~\cite{huang2021multiway} using their open-source implementations\footnote{\url{https://github.com/rabbityl/DeformationPyramid}}$^,$\footnote{\url{https://github.com/WanquanF/RMA-Net}}$^,$\footnote{\url{https://github.com/huangjh-pub/synorim}}. 
For each frame, we first downsample the depth map and obtain a point cloud according to the intrinsic camera parameters, then remove the background using the ground-truth segmentation masks provided by the dataset. 
Finally, we construct a triangular mesh from the point cloud according to their adjacency in the depth map, and extract the largest connected part.
We then use either the resulting mesh or its vertices as the input to a method depending on its requirement.
Since SyNoRiM requires 8192 points on each input shape, when the number of points on an input shape is different from 8192 we either downsample the points or add repeated points to achieve the required number.
To improve the performance of RMA-Net, we fine-tune the model on the 2120 data pairs from the training set of the DeepDeform dataset. For other learning-based methods, since they already test their performance on the DeepDeform dataset, we directly evaluate their performance using their pre-trained models. We perform the same processing on the training data as mentioned above for the testing data. 
The training and testing of the learning-based  methods are run on an NVIDIA GeForce RTX 3090 GPU.  
For both SVR-$\ell_0$ and our method, we set the sampling radius $R = 8\overline{l}$ for the deformation graph construction for better efficiency. 
For the optimization-based methods, we first perform a coarse alignment using the rigid registration method from~\cite{zhang2021fast} before running each method for non-rigid registration.
Tab.~\ref{Tab:deepdeform} shows the mean and median error measures and average computational time for each method, while Fig.~\ref{Fig:deepdeform} shows some example results from the methods. 
Overall, the optimization-based methods achieve lower registration errors than the learning-based methods, which is likely due to the capability of optimization-based methods to locally refine the alignment.
RMA-Net has worse accuracy than other learning-based methods, because its training does not consider partial overlaps between the input shapes.
Our method achieves the lowest mean value of $r_s$ and the third lowest median value of $r_s$ (and is only 3\% higher than the lowest median $r_s$ value), which shows the good accuracy of our method.
On the other hand, the learning-based methods are overall faster than the optimization-based methods, thanks to the efficiency of running a neural network model on the GPU. Nevertheless, our method is only slower than SyNoRiM and is significantly faster than other optimization-based methods.

\begin{figure}[t]
	\centering
	\includegraphics[width=\columnwidth]{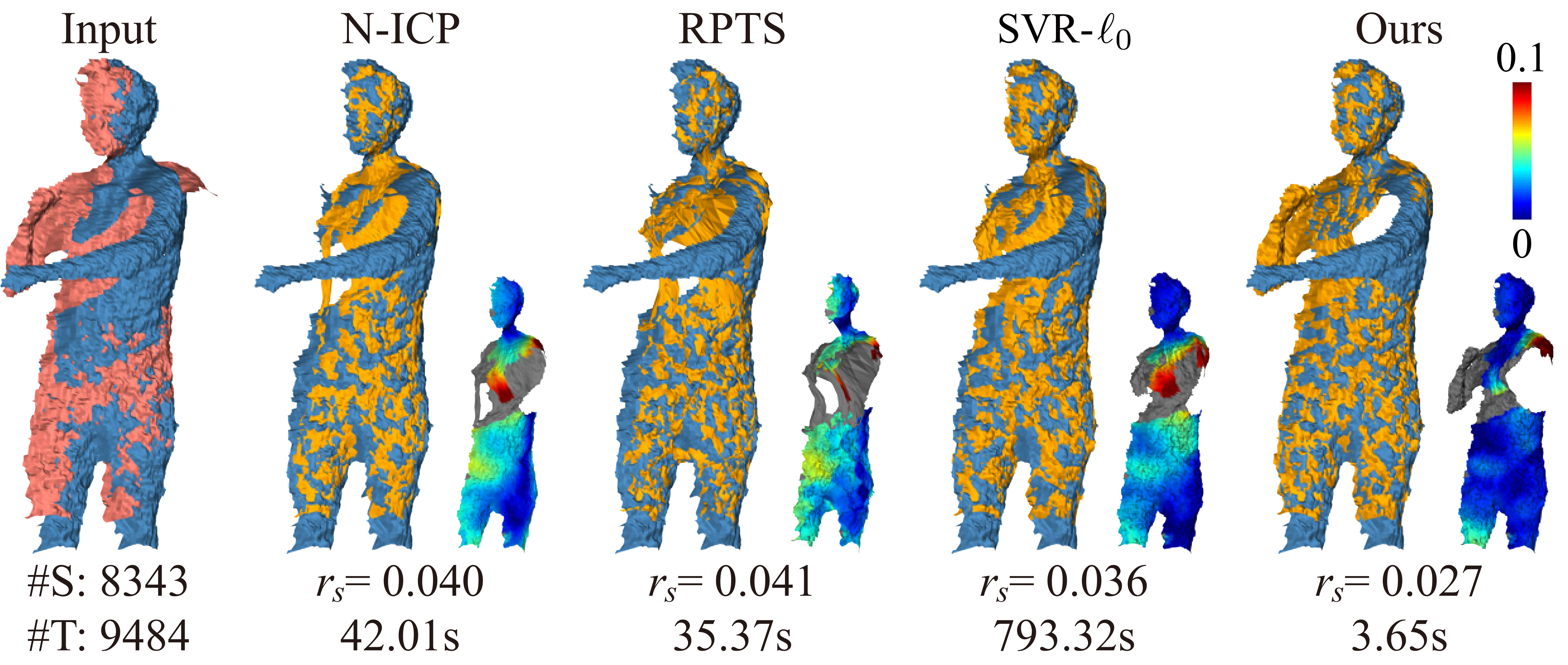}
	\caption{Comparison between different methods on a Raspberry Pi for DeepDeform dataset~\cite{bozic2020deepdeform}. We set $\alpha=10$ for N-ICP, $\alpha=100, \beta = 1$ for RPTS, $\alpha=1, \beta = 1, R=8\overline{l}$ for SVR-$\ell_0$, and $k_{\alpha}=10, k_{\beta}=1, R=8\overline{l}$ for our method. In the visualized error map, we mark the points without ground-truth scene flows in gray.}
	\label{Fig:deepdeform2}
\end{figure}

\begin{figure}[t]
	\centering
	\includegraphics[width=\columnwidth]{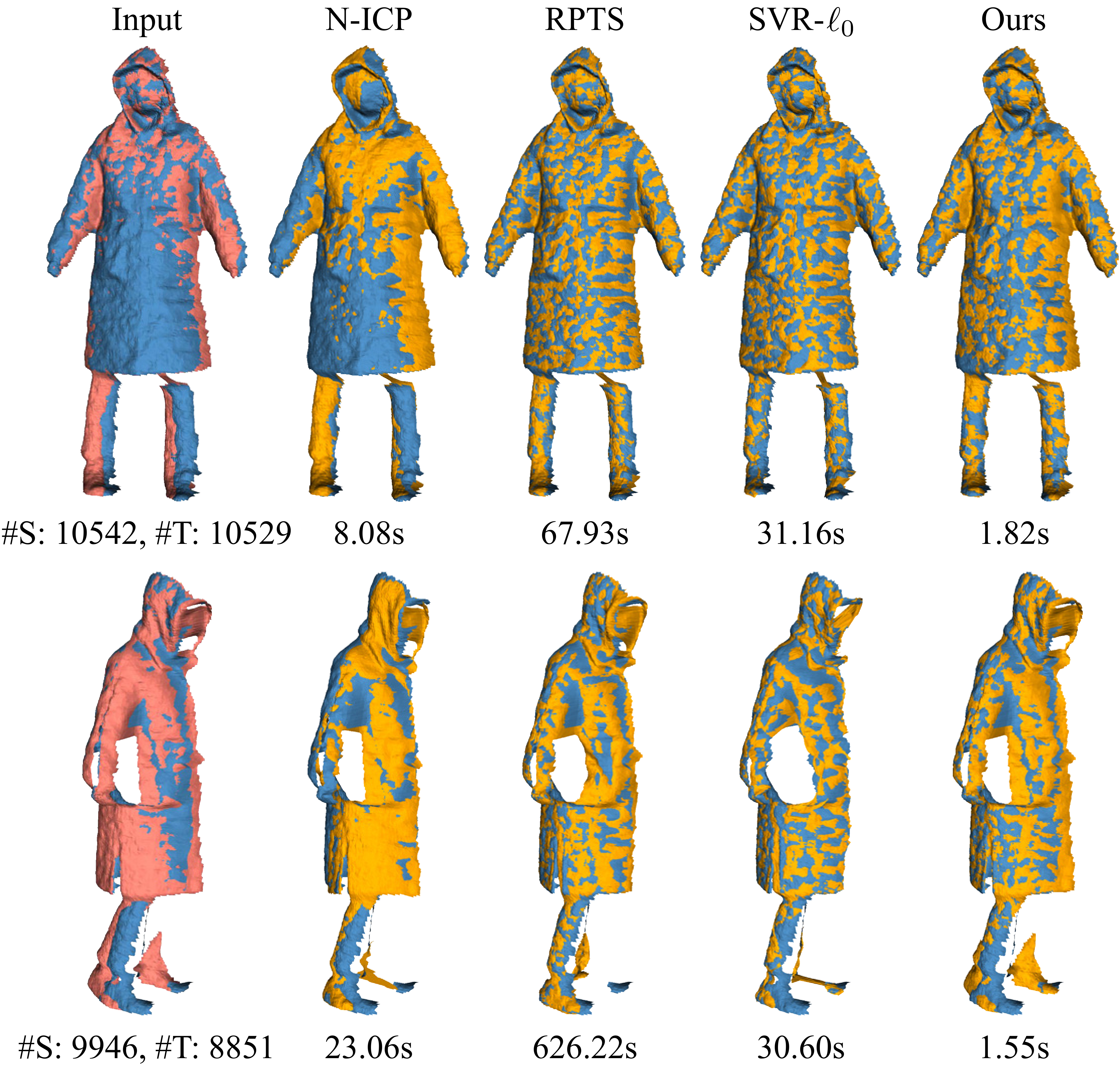}
	\caption{Comparison between different methods on a Raspberry Pi for real data collected with Kinect. We set $\alpha=100$ for N-ICP, $\alpha=10, \beta = 1$ for RPTS, $\alpha=1, \beta = 1$ for SVR-$\ell_0$, and $k_{\alpha}=10, k_{\beta}=100$ for our method.}
	\label{Fig:kinect2}
\end{figure}

\para{Performance on mobile devices}
Furthermore, due to the use of the deformation graph and our efficient solver, our method can be deployed on devices with limited compute capabilities.
To this end, we compare our method and other optimization-based methods on a Raspberry Pi 4B with 8GB of RAM and a quad-core Cortex-A72 at 1.5GHz running Ubuntu 20.04.
In Fig.~\ref{Fig:deepdeform2}, we show the comparison on a problem instance from the the DeepDeform dataset.
In Fig.~\ref{Fig:kinect2}, we compare the methods on data pairs obtained by scanning a self-rotating human subject with a Kinect camera; the scans are converted to triangle meshes for registration, by removing isolated points and connecting the remaining points according to their adjacency in the depth map.
Due to the lower compute power of Raspberry Pi, we relax the termination condition of our method to $\epsilon=3\times 10^{-4}$, which causes no notable change to the result while still being tighter than other methods. We can see that our method is significantly faster than other methods and achieves similar or better accuracy.

\section{Conclusion}

In this paper, we proposed a robust non-rigid registration model based on Welsch's function. Applying the
Welsch's function to the alignment term and the regularization term makes the formulation robust to noises and
partial overlaps. To efficiently solve this problem, we apply
majorization-minimization to transform the nonlinear and
non-convex problem into a sequence of simple sub-problems. To speed up the convergence, we regard the MM algorithm as a fixed-point iteration and use Anderson acceleration to accelerate the solution. Extensive experiments demonstrate the effectiveness of our method and its
efficiency compared to existing approaches.

\begin{figure}[t]
	\centering
	\includegraphics[width=\columnwidth]{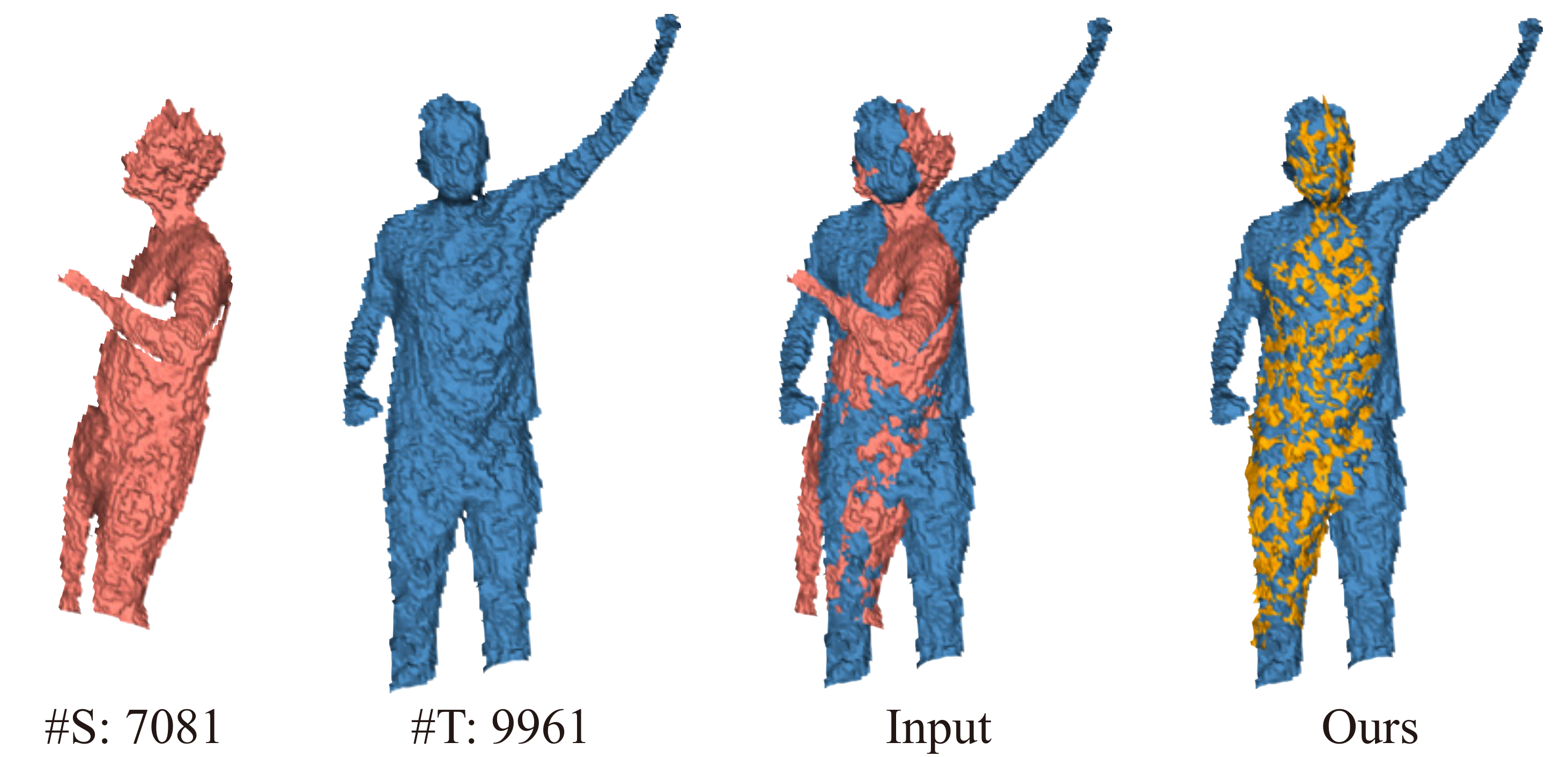}
	\caption{A failure case of our method on a problem instance from the DeepDeform dataset~\cite{bozic2020deepdeform}, where the significant difference between the source and target models leads to inaccurate alignment.}
	\label{Fig:failure_case}
\end{figure}

Although our method achieves good results on many examples, it still has some limitations. First, although our use of the robust metric helps to mitigate the impact of incorrect correspondence, our non-convex formulation still requires proper initialization to avoid converging to an undesirable local minimum. In particular, when there is a significant difference between the source and target shapes, the closest-point correspondence may be erroneous on a large part of the surface and lead to inaccurate alignment (see Fig.~\ref{Fig:failure_case} for an example of such failure cases). The issue can potentially be addressed by adopting a more sophisticated point correspondence method that also accounts for the local shape features.
Secondly, our deformation graph construction method relies on a global radius parameter that leads to an approximately uniform density of graph nodes.
Such a deformation graph may not be suitable for meshes with highly non-uniform samples. This can be improved using an adaptive radius parameter that depends on the local curvatures and sampling densities, which we will leave as future work.

\ifCLASSOPTIONcompsoc
  \section*{Acknowledgments}
\else
  \section*{Acknowledgment}
\fi

This work was supported by the National Natural Science Foundation of China (No. 62122071), the Youth Innovation Promotion Association CAS (No. 2018495), and the Fundamental Research Funds for the Central Universities (No. WK3470000021).

\bibliographystyle{IEEEtran}

\bibliography{FastNrr}

\ifCLASSOPTIONcaptionsoff
  \newpage
\fi

\begin{IEEEbiography}[{\includegraphics[width=1in]{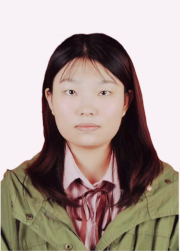}}]{Yuxin Yao} is currently working toward a PhD degree in the School of Mathematical Sciences, University of Science and Technology of China. Her research interests include computer vision, computer graphics and numerical optimization.
\end{IEEEbiography}

\begin{IEEEbiography}[{\includegraphics[width=1in]{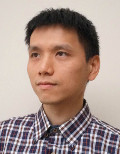}}]{Bailin Deng}
	is a Senior Lecturer in the School of Computer Science and Informatics at Cardiff University. He received the BEng degree in computer software (2005) and the MSc degree in computer science (2008) from Tsinghua University (China), and the PhD degree in technical mathematics (2011) from Vienna University of Technology (Austria). His research interests include geometry processing, numerical optimization, computational design, and digital fabrication. He is a member of the IEEE.
\end{IEEEbiography}

\begin{IEEEbiography}[{\includegraphics[width=1in]{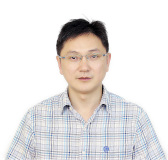}}]{Weiwei Xu}
is a researcher with the State Key Lab of CAD \& CG, College of Computer Science, Zhejiang University, awardee of NSFC Excellent Young Scholars Program in 2013. His main research interests include the digital geometry processing, physical simulation, computer vision and virtual reality. He has published around 70 papers on international graphics journals and conferences, including 20 papers on ACM TOG. He is a member of the IEEE and ACM.
\end{IEEEbiography}

\begin{IEEEbiography}[{\includegraphics[width=1in]{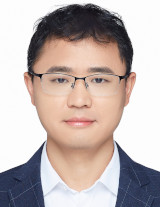}}]{Juyong Zhang}
is a professor in the School of Mathematical Sciences at University of Science and Technology of China. He received the BS degree from the University of Science and Technology of China in 2006, and the PhD degree from Nanyang Technological University, Singapore. His research interests include computer graphics, computer vision, and numerical optimization. He is an associate editor of IEEE Transactions on Multimedia and The Visual Computer.

\end{IEEEbiography}

\vfill

\end{document}